\documentclass{article}

 \usepackage[main, final]{neurips_2026}
\usepackage[utf8]{inputenc} % allow utf-8 input
\usepackage[T1]{fontenc}    % use 8-bit T1 fonts
\usepackage{hyperref}       % hyperlinks
\usepackage{url}            % simple URL typesetting
\usepackage{booktabs}       % professional-quality tables
\usepackage{amsfonts}       % blackboard math symbols
\usepackage{nicefrac}       % compact symbols for 1/2, etc.
\usepackage{microtype}      % microtypography
\usepackage{xcolor}         % colors
\usepackage{graphicx}

\usepackage{amsmath}
\usepackage{cleveref}
\usepackage{multirow}
\usepackage{caption}
\newtheorem{proposition}{Proposition}
\usepackage{wrapfig}
\usepackage{subcaption}
\hypersetup{
    colorlinks=true,
    linkcolor=red,
    citecolor=blue,
    urlcolor=blue
}
\title{Sparse All-Layer Connector for Domain Generalised Semantic Segmentation}

\author{%
  Ruoyu Guo\thanks{Equal contribution.} \quad
  Xin Kun Lin\footnotemark[1] \quad
  Maurice Pagnucco \quad
  Yang Song \\
  School of Computer Science and Engineering, University of New South Wales\\
  \texttt{ruoyu.guo@student.unsw.edu.au} \\
  \texttt{\{xin\_kun.lin, morri, yang.song1\}@unsw.edu.au} \\
}

\begin{document}

\maketitle

\begin{abstract}
  Domain generalised semantic segmentation (DGSS) has recently benefited from combining multiple foundation models, such as vision-language and vision foundation models, to leverage their diverse representations for generalisation. Existing multi-model approaches share a common design choice: cross-model fusion is restricted to depth-aligned layers, implicitly assuming that mutually beneficial information resides at the same depth across models. However, foundation models pretrained under different objectives may develop distinct representation hierarchies, making the optimality of depth-aligned fusion questionable. Moreover, we observe that the foundation models exhibit different depth-wise domain sensitivity. Motivated by these observations, we propose a Sparse All-Layer Connector (SALC) that enables each layer to access information from preceding layers of another model. As the number of accessible layers grows with depth, dense aggregation may hurt generalisation. SALC therefore learns to adaptively select a sparse subset of informative layers. We further introduce a candidate dropout regularisation that strengthens sparsity and encourages SALC to explore diverse layers, leading to more robust selection. Across four foundation model combinations and three DGSS evaluation settings, the proposed design demonstrates stronger generalisation than conventional depth-aligned fusion. The code is available at \url{https://github.com/RuoyuGuo/SALC}.
\end{abstract}

\section{Introduction} 
\label{sec:intro}
Domain generalised semantic segmentation (DGSS) seeks to learn a segmentation model that performs robustly on arbitrary unseen target domains using only source domain data. To extract robust features, prior work has explored a range of strategies~\cite{choi2021robustnet,pan2019switchable}, while more recent work leverages foundation models for their strong generalisability~\cite{rein}. Despite these advances, relying on a single foundation model limits the diversity of representations, as each model is bounded by its pretrained knowledge. This has motivated a growing body of work that integrates multiple foundation models to learn heterogeneous representations for DGSS, such as vision-language models (VLMs), depth estimation foundation models (DFMs), and vision foundation models (VFMs)~\cite{mfuser,chen2026leveraging,Chen_2025_ICCV,pak2024textual,yang2024depth}.

In practice, existing multi-model approaches~\cite{mfuser,chen2026leveraging,Chen_2025_ICCV} adopt depth-aligned fusion as a natural default: a learnable connector (e.g., linear or cross-attention layers) is placed at the $l$-th layer of one foundation model to fuse features from the $l$-th layer of the other. Previous studies have found that this approach can fuse invariant features or visually grounded features into a stronger representation~\cite{chen2026leveraging}. However, models pretrained under different objectives may process information along distinct hierarchies~\cite{cka3,sun2025transformer} and prior findings have shown that foundation models exhibit different transferability across layers~\cite{skean2025layer,odonnat2026layer}. In such cases, it is unclear whether depth-aligned pairing remains the most effective choice.

\begin{figure}[t]
    \centering
    \includegraphics[width=1\textwidth]{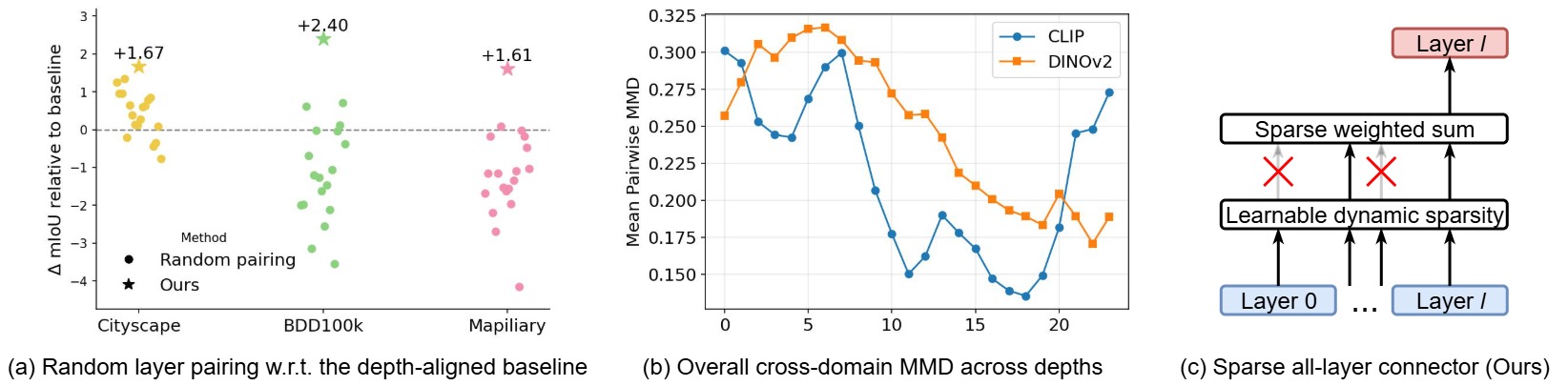}
    \caption{(a) The model is trained on GTAV and evaluated on Cityscapes, BDD100K, and Mapillary. The random pairing is sampled independently for DINOv2 and CLIP  using over 40 seeds. (b) Layer-wise cross-domain behaviour measured by MMD.  We report the average pairwise MMD over all six domain pairs of the four 
domains. (c) Architecture of our sparse all-layer fusion.}
    \label{fig:tesear}
\end{figure}

To examine this empirically, we study a representative combination~\cite{jiang2023clip,mfuser,tong2024eyes}, a VLM (CLIP~\cite{clip}) and a VFM (DINOv2~\cite{oquab2023dinov2}). We compare the depth-aligned baseline with a variant that randomly pairs layers across the two models. As shown in \Cref{fig:tesear} (a), random pairing yields mixed results relative to depth-aligned fusion, suggesting that the effectiveness is sensitive to layer matching. We further analyse layer-wise cross-domain behaviour using Maximum Mean Discrepancy (MMD). \Cref{fig:tesear} (b) shows that CLIP and DINOv2 exhibit markedly different depth-wise patterns: CLIP reaches its lowest at middle layers and increases toward the end, while DINOv2 decreases monotonically with depth. These observations motivate us to fuse features between models in a flexible cross-depth manner. In this paper, for each layer in one model, which we refer to as a \textit{query layer}, we allow it to aggregate information from multiple layers of the other model, which we refer to as \textit{candidate layers}.

However, this flexible design poses two critical challenges. First, the number of candidate layers grows with depth, and the informativeness of each candidate layer varies across tokens, which complicates effective fusion. Second, dense aggregation across all candidate layers can introduce irrelevant or domain-specific features, which may hurt generalisation. Although some existing methods~\cite{cao2024mmfuser,shen2024mome}  propose various multi-modal fusion strategies, they typically utilise cross-attention and TopK routing, which are insufficient here. This is because cross-attention uses softmax, which assigns non-zero weights to every candidate layer and may retain irrelevant features, while TopK routing enforces sparsity but with a fixed K that cannot adapt across depths and tokens.

% To address these, 

% Some existing approaches such as 

% how to identify which candidate layers are most informative and determine how many should be activated; and (2) how to make this selection mechanism robust under domain generalisation. 

% In this work, we explore the cross-layer adapter to aggregate the multi-model information for DGSS. We adopt a similar framework as prior works~\cite{mfuser,Chen_2025_ICCV}, where two foundation models extract features, and a mask2former~\cite{cheng2022masked} decoder is used to produce the segmentation map. Moreover, an adapter is inserted after each layer to fuse features from foundation models. 

To this end, we propose the \textbf{Sparse All-Layer Connector (SALC)} for cross-model fusion in DGSS. Different from existing methods that restrict fusion to depth-aligned
layers, SALC allows each query layer to access an adaptive subset of all preceding layers up to the same depth in the other model, so that only the most relevant features are fused.  To learn adaptive selection, we adopt Sparsegen~\cite{laha2018controllable}, which projects predicted layer scores into a sparse weight distribution over candidate layers. Unlike softmax or TopK routing, Sparsegen can control the active candidate layers through a sparsity factor rather than a manually defined number. Building on this, we predict a token-wise sparsity factor using the candidate layer at the same depth, thereby enabling adaptivity at both the \textit{token} and \textit{layer} levels.

Moreover, we propose a candidate dropout regularisation that perturbs over-selected layers during training, guiding SALC toward more robust and discriminative selection. We explore two instantiations: \emph{uniform drop}, where the same drop rate is applied for all layers and \emph{frequency-based drop}, where a higher drop rate is applied for over-used layers. We evaluate SALC on five foundation models spanning VLMs (CLIP, SigLIP~\cite{zhai2023sigmoid}), VFMs (DINOv2, SAM~\cite{kirillov2023segment}), and a depth model (DepthAnythingV2~\cite{yang2024depthv2})
 across three evaluation settings, showing competitive performance compared to depth-aligned fusion. Our main contributions are summarised as follows:
\begin{itemize}
    \item We propose SALC, a multi-model DGSS framework that aggregates cross-layer features from different foundation models through a sparse selection mechanism. The selection is adaptively controlled for each layer and token through a learnable sparsity factor.
    \item To achieve more robust layer selection, we further introduce a candidate dropout regularisation that strengthens sparsity and exploration of informative layers.
    \item We demonstrate the effectiveness of SALC and provide an in-depth analysis of how cross-layer selection benefits domain generalisation. We also achieve competitive performance on three types of foundation models across three DGSS evaluation settings.
\end{itemize}

\section{Related work}

% \subsection{Style}
\textbf{Domain generalised semantic segmentation (DGSS).} \quad DGSS requires the segmentation model to remain robust when encountering unseen images~\cite{ding2023hgformer,hoyer2022hrda,kim2022pin,liuncertainty,luqprompt}. Classic approaches mainly follow three directions, including data augmentation~\cite{yue2019domain,lee2022wildnet,kim2023texture,NEURIPS2022_023d94f4,huang2021fsdr,chattopadhyay2023pasta,jiang2023domain,jia2024dginstyle,benigmim2024collaborating,fahes2024simple,zhao2022style,wu2022siamdoge}, learning invariant representation~\cite{ahn2024style,huang2023style,xu2022dirl,huang2021fsdr,choi2021robustnet,bi2024learningAAAI} and feature normalisation~\cite{pan2018two,pan2019switchable,peng2022semantic,chen2025exploring}. Despite their effectiveness, these methods rely on models trained with less data. Recent methods adapt foundation models to DGSS through parameter-efficient fine-tuning~\cite{rein,hummer2023vltseg} or selectively tuning domain-sensitive parameters and features~\cite{bi2024learning, yun2025soma,yi2024learning,zhao2025fishertune,bi2025nightadapter}. To extract more diverse cues, several methods fuse features from multiple foundation models, such as language~\cite{pak2024textual,mfuser,niu2025exploring,jeon2025exploiting,lin2026hierarchical}, depth map~\cite{Chen_2025_ICCV,chen2026leveraging}, and high-resolution images~\cite{unleashing}. Unlike prior multi-model DGSS methods that primarily explore different model combinations, we improve generalisation through the fusion design itself. We propose flexible cross-depth layer fusion together with a frequency-based dropout regularisation that encourages SALC to discover more robust features.

\textbf{Multi-model feature fusion.} \quad 
Multi-model feature fusion aims to exploit complementary representations from different encoders or modalities~\cite{lin2025multi}. In multimodal large language models, one line of work enhances visual perception by combining visual tokens from multiple encoders, resolutions, or expert branches~\cite{lin2023sphinx,kar2024brave,tong2024cambrian,li2025mini,shen2024mome,shieagle,lin2025multi}. Another line aggregates multi-layer features before feeding into the language model with depth-aligned or dense connectors~\cite{luo2025feast,meng2024deepstack,cao2024mmfuser,jiang2023clip,chen2024lion,chen2024evlm,NEURIPS2024_3a10c465}. In DGSS, recent multi-model methods fuse complementary cues from VFMs, VLMs, and depth foundation models, but their cross-model interactions are typically established between depth-aligned layers~\cite{mfuser,chen2026leveraging,Chen_2025_ICCV}. 
% These designs improve the diversity of visual cues but adopt depth-aligned fusion as a default choice. This assumption is questionable under domain shift, where different layers exhibit different transferability across models. 
Unlike these works, SALC tackles cross-model layer assignment as a learnable problem: for each query layer, it adaptively selects a sparse subset of candidate layers. 
% This makes cross-model fusion more selective and better suited to DGSS, where irrelevant or domain-specific candidate features may hurt generalisation.

\section{Method}
 
\subsection{Overview}

\textbf{Problem formulation.} \quad We address the DGSS problem, where a model is trained on a 
labelled source domain $\mathcal{D}_s$ and evaluated on unseen target domains $\mathcal{D}_t$ whose 
samples are inaccessible during training. We use 
$\mathcal{D}_s \rightarrow \mathcal{D}_t$ to denote training on 
$\mathcal{D}_s$ and evaluating on $\mathcal{D}_t$.

\begin{figure}[t]
    \centering
    \includegraphics[width=1\textwidth]{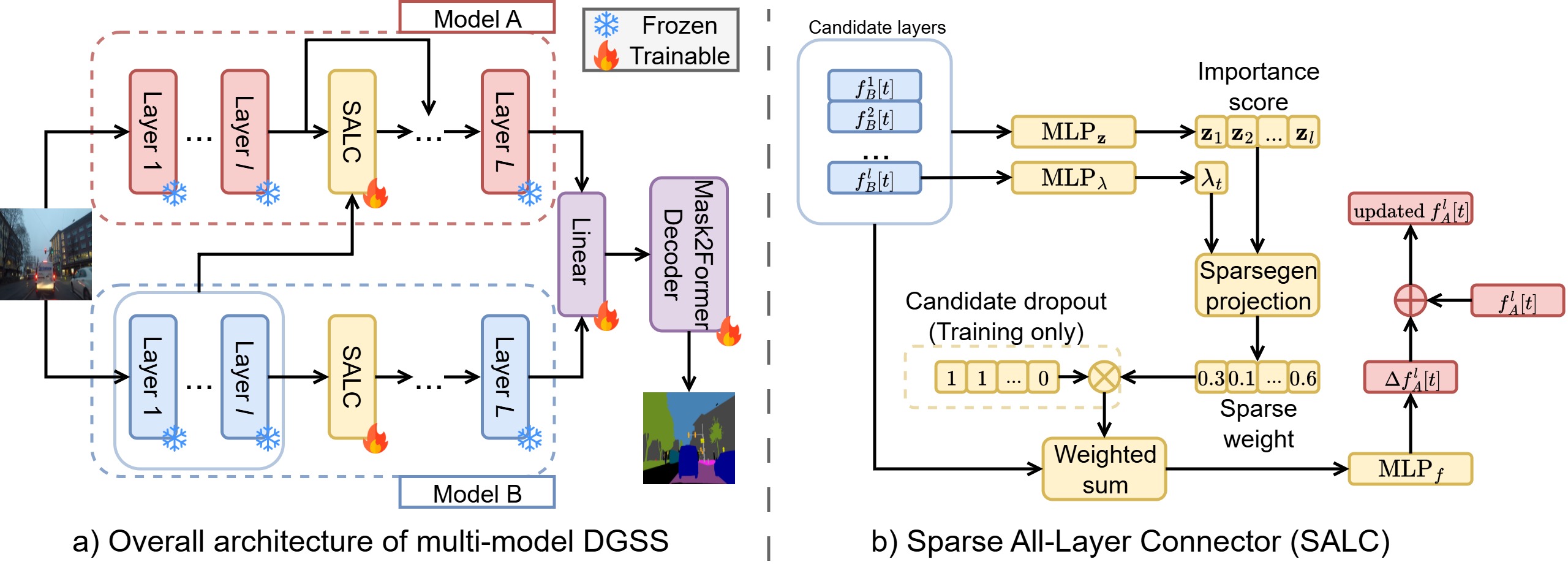}
    \caption{(a) Two models extract features from the same image in parallel. SALC fuses multi-layer features from one model into the other. The final segmentation map is predicted using multi-scale features of two models. (b)  SALC selectively fuses features from all candidate layers, controlled by a learnable sparsity factor. We also propose a dropout regularisation to improve robustness.}
    \label{fig:framework}
\end{figure}

\textbf{Overall architecture.}\quad 
As illustrated in Figure~\ref{fig:framework}, our framework consists 
of two frozen encoders $\mathcal{E}_A$ and $\mathcal{E}_B$ 
pretrained under different objectives and a segmentation decoder. The core component of our framework is a set of SALCs attached after each layer, allowing each \emph{query layer} to access all preceding \emph{candidate layers} while activating only a sparse subset for aggregation. To improve the robustness of selection, we further propose a frequency-based regularisation. Given an input RGB image, the two encoders extract features in parallel. The last-layer features from both encoders are then concatenated and projected by a linear layer into a unified representation, which is fed into a 
Mask2Former~\cite{cheng2022masked} decoder to produce the final segmentation map.

\subsection{Sparse all-layer connector}
\label{sec:salc}

\textbf{Revisiting depth-aligned fusion.} \quad % We write $f_A^l[t] \in \mathbb{R}^D$ for the feature of the $t$-th token at layer $l$ from $\mathcal{E}_A$, with $t \in \{1,...,T\}$. 
% \textbf{Connector formulation.}\quad 
In prior multi-model DGSS works~\cite{mfuser, chen2026leveraging,Chen_2025_ICCV}, given an input image $x$, it is fed into two encoders $\mathcal{E}_A$ and $\mathcal{E}_B$ with $L$ 
transformer layers each. We denote the output of their $l$-th 
layer as $f_A^l, f_B^l\in \mathbb{R}^{T \times D}$, where $T$ and $D$ are the number of tokens and feature dimensions. A depth-aligned connector $\mathcal{C}_l$ is employed to fuse 
features from the same layer of each encoder and residually 
updates the query encoder: $ f_A^l \leftarrow f_A^l + \Delta f_A^l,$ where $\Delta f_A^l = \mathcal{C}_l(f_A^l, f_B^l) \in \mathbb{R}^{T \times D}.$  Such depth-aligned pairing, however, is not necessarily optimal.

Here, we propose SALC that jointly learns candidate 
weights and adaptively controls the number of active 
candidate layers. Unlike prior 
work~\cite{mfuser, yang2024depth, chen2026leveraging}, SALC produces an update signal $\Delta f_A^l \in \mathbb{R}^{T\times D}$ that aggregates information from \emph{all} accessible 
candidates $\{f_B^1, \ldots, f_B^l\}$ rather than $f_B^l$ alone. Moreover, it learns token-specific sparsity levels for each layer.  Let $[t]$ denote indexing at the $t$-th token position, 
we describe how $\Delta f_A^l[t]$ is computed in three steps: (1) predicting importance scores for each candidate layer, (2) projecting them into a sparse weight distribution, and (3) aggregating candidates with the resulting weights. This pipeline is applied independently at each token position and symmetrically to $\mathcal{E}_B$. 

\textbf{Importance scoring.}\quad For the \textit{l}-th query layer, we collect the outputs at position $t$ from all accessible candidate layers as $ \mathbf{F}_t=[f_B^1[t], \ldots, f_B^l[t]] \in \mathbb{R}^{l \times D} $.  We then predict importance scores for $l$ layers using an MLP, while an 
RMSNorm is applied first to mitigate the magnitude 
variations across depths:
\begin{equation}
    \mathbf{z}_t = \text{MLP}_z\big(\text{RMSNorm}(\mathbf{F}_t)\big) 
    \in \mathbb{R}^{l},
\label{eq:score}    
\end{equation}

\textbf{Sparse weight projection.}\quad To adaptively select useful layers from $\mathbf{F}_t$ for aggregation, we leverage 
Sparsegen~\cite{laha2018controllable} to map $\mathbf{z}_t$ into a sparse weight distribution $\mathbf{w}_t\in \mathbb{R}^l$.
Specifically, Sparsegen activates only the candidate layers whose importance scores exceed a threshold $\tau_t\in \mathbb{R}$.  The threshold is jointly determined by $\mathbf{z}_t$ and a sparsity factor $\lambda_t < 1$. Intuitively, a larger $\lambda_t$ results in a higher threshold, leading SALC to concentrate on fewer layers, while a smaller $\lambda_t$  encourages SALC to aggregate more candidate layers. Formally, the sparse weights are computed as:
\begin{equation}
    \mathbf{w}_t = \max\left(\frac{\mathbf{z}_t - \tau_t}{1 - \lambda_t}, 0\right),
    \quad \text{where} \quad
    \tau_t = 
    \frac{\sum_{k \in \mathcal{S}_t} \mathbf{z}_t[k] - 1 + \lambda_t}{|\mathcal{S}_t|}.
\label{eq:weight}
\end{equation}

% where $\tau_t$ is determined as:

% \begin{equation}
%     \tau_t = \frac{\sum_{k \in \mathcal{S}} \mathbf{z}_t[k] - 1 + \lambda_t}{|\mathcal{S}_t|},
% \label{eq:tau}   
% \end{equation}

The active set $\mathcal{S}_t \subseteq \{1,...,l\}$  contains the top-ranked candidate layers by their importance scores (we provide the procedure for computing the size of $\mathcal{S}_t$ in Appendix~\ref{app:sparsegen}).  

Since semantic segmentation requires per-pixel predictions and spatially distinct tokens may benefit from different sparsity levels, we predict token-wise $\lambda_t$ using $f_B^l$. This is because $f_B^l$ provides a summarised view of all preceding candidate layers through residual updates, offering an effective signal to estimate the informativeness of previous features and determine the sparsity. Predicting $\lambda_t$ from $f_B^l$ also makes the sparsity input-dependent, allowing $\lambda_t$ to vary with the content of each image rather than being fixed across inputs. We use an MLP for prediction:
\begin{equation}
    \lambda_t = \text{MLP}_\lambda (f_B^l[t]).
\label{eq:lambda}    
\end{equation}

\textbf{Candidate aggregation.}\quad We then 
aggregate $\mathbf{F}_t$ using the sparse weights 
$\mathbf{w}_t$, followed by an MLP that projects the result into 
the feature space of $\mathcal{E}_A$:
\begin{equation}
    \Delta f_A^l[t] = \text{MLP}_f\Big(\mathbf{w}_t^\top \cdot \mathbf{F}_t\Big) \in \mathbb{R}^{D}.
\label{eq:delta}    
\end{equation}
 We use shared $\text{MLP}_z$, $\text{MLP}_\lambda$ and $\text{MLP}_f$ for all tokens in each SALC to reduce complexity. Across query layers, SALCs are independently parameterised and do not share weights.
%to finally obtain $\Delta f^l_A$.

\subsection{Candidate dropout regularisation}
The effectiveness of SALC relies on its ability to make input-dependent selection of candidate layers. However, during training, we observe that at certain query layers, the selection weights spread too broadly across candidate layers, suggesting the module fails to make decisive choices at these layers. We thus propose a regularisation that drops accessible candidate layers, encouraging more decisive selection at every layer. We explore two instantiations: \emph{uniform dropout} and 
\emph{frequency-based dropout}. The regularisation is applied only during training and does not incur inference costs.

At each training step, we sample a binary mask 
$\mathbf{M} \in \{0, 1\}^{l}$, where 
$\mathbf{M}[i] = 0$ ($1\le i \le l$) indicates dropping the $i$-th candidate. 
We stack the per-token weights from \Cref{eq:weight} into 
$\mathbf{W} \in \mathbb{R}^{T \times l}$, multiply $\mathbf{W}$ by $\mathbf{M}$ to drop candidate layers, and 
renormalise each row so that the weights sum to 1 for every token. The 
renormalised $\tilde{\mathbf{W}}[i]\in \mathbb{R}^l$ then replaces $\mathbf{w}$ in \Cref{eq:delta} for aggregation. 

For sampling $\mathbf{M}$, we first explore \emph{uniform 
dropout}, where each candidate layer is independently dropped with 
a fixed probability $p$. This approach treats all candidate layers equally. We further explore \emph{frequency-based dropout}, where the drop probability $p$ of each candidate layer is scaled by its historical selection frequency, giving dominant candidate layers a higher chance of being dropped. Specifically, we maintain two running statistics via 
exponential moving average (EMA). The first, $\bar{\mathbf{u}} \in \mathbb{R}^l$, tracks 
the average weight assigned to $l$ candidate layers, obtained by 
averaging $\mathbf{W}$ over the token dimension. The second, $\bar{N} \in \mathbb{R}$, 
tracks the average number of active candidates (whose $\mathbf{W}>0$) per token. We set the target usage to $\rho=1/\bar{N}$. 
The drop probability for candidate layer $i$ is:
\begin{equation}
    p_i = p_{\max} \cdot \left[ \text{max}\left( \frac{\bar{\mathbf{u}}_i - \rho}{1 - \rho}, 0 \right)\right]^{\gamma},
\end{equation}
where $p_{\max}$ caps the drop rate and $\gamma$ controls the 
penalty sharpness. Mask entries are sampled as $\mathbf{M}[i]
\sim \text{Bernoulli}(1 - p_i)$ in both instantiations.

Intuitively, \emph{frequency-based dropout} penalises only the candidates whose usage exceeds the target $\rho$, such that candidates with usage below $\rho$ receive zero drop probability. With $\gamma > 1$, the drop rate grows sharply with overuse, concentrating the regularisation on the most over-used candidates. We apply this regularisation independently within each SALC, 
with $\bar{\mathbf{u}}$ and $\bar{N}$ tracked per-SALC to capture the 
selection pattern of each query layer.

\subsection{Training objectives}
Following prior multi-model DGSS work~\cite{mfuser}, we train SALC using the standard segmentation losses from Mask2Former~\cite{cheng2022masked}: $\mathcal{L}_{total}= \alpha_{bce}\mathcal{L}_{bce} + \alpha_{dice}\mathcal{L}_{dice} + \alpha_{cls}\mathcal{L}_{cls}$,  where $\mathcal{L}_{bce}, \mathcal{L}_{dice}, \text{ and }\mathcal{L}_{cls}$ represent the binary cross-entropy loss, the dice loss for the predicted masks, and the cross-entropy loss for each queried proposal, respectively.

% $L * T $
% We explore two instantiations of $\mathbf{M}$:

% according to a predefined drop rate $p_{rand}$.

% \textbf{Uniform dropout.}  \quad We first utilise a vanilla strategy to sample $\mathbf{M}$, while

% \textbf{Frequency-based dropout.} \quad 

\section{Experiments}
\subsection{Datasets \& evaluation protocols}
\textbf{Datasets.}\quad We evaluate SALC on five driving-scene semantic segmentation datasets that share 19 common classes. Specifically, GTAV (G)~\cite{richter2016playing} and SYNTHIA (S)~\cite{ros2016synthia} are two synthetic datasets, containing 24,966 and 9,400 training images, respectively. Cityscapes (C)~\cite{cordts2016cityscapes} consists of 2,975 training and 500 validation images. BDD100K (B)~\cite{yu2020bdd100k} and Mapillary (M)~\cite{neuhold2017mapillary} include 1,000 and 2,000 validation images, respectively. 

% These three real-world datasets capture diverse scenes from a variety of global cities. 

% We additionally utilise ACDC~\cite{sakaridis2021acdc}, offering 406 validation images captured under adverse conditions (night, snow, fog, and rain). 

\textbf{Evaluation.}\quad Following the evaluation protocols of existing DGSS methods~\cite{rein,bi2024learning,pan2018two}, we evaluate SALC on three evaluation settings with the mIoU metric: (1) G$\rightarrow$\{C, B, M\}, (2) S$\rightarrow$\{C, B, M\}, and (3) C$\rightarrow$\{B, M\}.
% , and (4) C$\rightarrow$ACDC. 

\textbf{Implementation details.}\quad Following existing multi-model DGSS methods~\cite{mfuser,Chen_2025_ICCV,chen2026leveraging}, we evaluate SALC on two VFMs (SAM~\cite{kirillov2023segment}, DINOv2~\cite{oquab2023dinov2}), two VLMs (CLIP~\cite{clip}, SigLIP~\cite{zhai2023sigmoid}), and one DFM (DepthAnythingV2~\cite{yang2024depthv2}). This results in four combinations of $\mathcal{E}_A$ and $\mathcal{E}_B$: DINOv2-CLIP, DINOv2-SigLIP, SAM-CLIP, and DINOv2-DepthAnythingV2. To ensure fair comparison, for the first three combinations, we use the framework of MFuser~\cite{mfuser}. For DINOv2-DepthAnythingV2, we use the framework of DepthForge~\cite{Chen_2025_ICCV}. More details are shown in Appendix~\ref{app:details}.

%More details are shown in Appendix~\ref{app:details}.

\subsection{Comparison with State-of-the-art DGSS methods}
We compare SALC with various DGSS methods: (1) classic methods, including SAN-SAW~\cite{peng2022semantic},  AdvStyle~\cite{NEURIPS2022_023d94f4}, and CMFormer~\cite{bi2024learningAAAI}; (2) Single foundation model methods, including tqdm~\cite{pak2024textual}, Rein~\cite{rein}, SET~\cite{yi2024learning}, FADA~\cite{bi2024learning}, CLOUDS~\cite{benigmim2024collaborating} and SoMA~\cite{yun2025soma}; and (3) Multi foundation model methods, including MFuser~\cite{mfuser}, DepthForge~\cite{Chen_2025_ICCV} and Vireo~\cite{chen2026leveraging}.

\begin{figure}[t]
    \centering
    \includegraphics[width=1\textwidth]{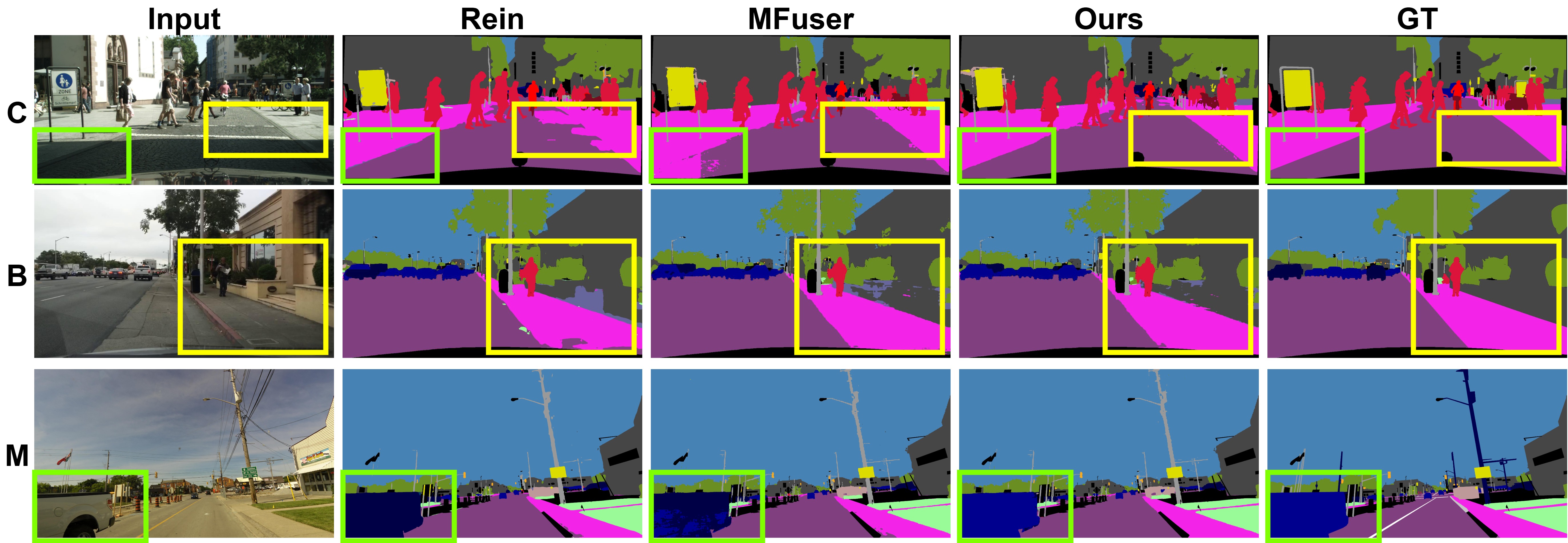}
    \caption{Qualitative comparison with Rein~\cite{rein} and MFuser~\cite{mfuser} under the G$\rightarrow$\{C, B, M\} setting.}
    \label{fig:maincompare}
\end{figure}

We report results on \textbf{G$\rightarrow$\{C, B, M\}}, \textbf{S$\rightarrow$\{C, B, M\}}, and \textbf{C$\rightarrow$\{B, M}\} in \Cref{tab:main_results}. SALC achieves competitive performance across all three settings, outperforming both single-foundation-model and depth-aligned multi-model baselines on most backbone configurations. On G$\rightarrow
$\{C, B, M\}, SALC achieves an average of 69.40 with DINOv2-CLIP, surpassing the previous depth-aligned baseline MFuser by 1.58 mIoU and the strongest single-foundation-model method (SoMA) by 1.13 mIoU. The improvement is larger on the challenging S$\rightarrow$\{C, B, M\}, where SALC reaches 52.99 with DINOv2-CLIP, surpassing MFuser by +2.65.  Notably, while multi-model baselines like DepthForge and MFuser sometimes underperform single-encoder CLOUDS on this setting, indicating that naive depth-aligned fusion may hurt generalisation under large domain shifts. SALC's effectiveness holds across multiple backbone combinations, reaching the largest single-domain gain of +4.16 mIoU over DepthForge on G$\rightarrow$C. We observe that SALC's improvements on SAM-CLIP are smaller and occasionally slightly negative. We provide further analysis in Appendix~\ref{app:limit}. We also provide qualitative comparisons in \Cref{fig:maincompare}.

\begin{table*}[t]
\centering
\small
\caption{Comparison with SOTA. * denotes re-implementation with official code. \textbf{Bold} and \underline{underline} denote the best and second-best. Depth denotes DepthAnythingV2. SYNTHIA has only 16 of the 19 evaluation classes in other datasets and we follow \cite{mfuser} to report the result over 19 classes, while some methods may report the result over 16 classes.}
\label{tab:main_results}
\resizebox{\textwidth}{!}{
\begin{tabular}{llccccccccccc}
\toprule
\multirow{2}{*}{Method}
& \multirow{2}{*}{Backbone}
& \multicolumn{4}{c}{Trained on GTAV (G)}
& \multicolumn{4}{c}{Trained on SYNTHIA (S)}
& \multicolumn{3}{c}{Trained on Cityscapes (C)} \\
\cmidrule(lr){3-6}
\cmidrule(lr){7-10}
\cmidrule(lr){11-13}
&
& $\rightarrow$C & $\rightarrow$B & $\rightarrow$M & Avg.
& $\rightarrow$C & $\rightarrow$B & $\rightarrow$M & Avg.
& $\rightarrow$B & $\rightarrow$M & Avg. \\
\midrule

\multicolumn{13}{l}{\textit{Classic DGSS}} \\
SAN-SAW~\cite{peng2022semantic} & ResNet-101 & 45.33 & 41.18 & 40.77 & 42.43 & 40.87 & 35.98 & 37.26 & 38.04 & 54.73 & 61.27 & 58.00 \\

AdvStyle~\cite{NEURIPS2022_023d94f4} & ResNet-101 & 39.52 & 36.39 & 36.10 & 37.34 & 37.59 & 27.45 & 31.76 & 32.27 & -- & -- & -- \\
CMFormer~\cite{bi2024learningAAAI}
& Swin-L & 55.31 & 49.91 & 60.09 & 55.10 & 44.59 & 33.44 & 43.25 & 40.43 & 59.27 & 71.10 & 65.19 \\
\midrule

\multicolumn{13}{l}{\textit{Single foundation model}} \\
Rein~\cite{rein} & DINOv2 & 66.40 & 60.40 & 66.10 & 64.30 & 48.59 & 44.42 & 48.64 & 47.22 & 65.00 & 72.30 & 68.65 \\
FADA~\cite{bi2024learning} & DINOv2 & 68.23 & 61.94 & 68.09 & 66.09
& 50.04 & 45.83 & 49.86 & 48.58
& 65.12 & 75.86 & 70.49 \\
SET~\cite{yi2024learning} & DINOv2& 68.06 & 61.64 & 67.68 & 65.79 & 49.65 & 45.45 & 49.45 & 48.18 & 65.07 & 75.67 & 70.37 \\
SoMA~\cite{yun2025soma} & DINOv2&  71.82 & 61.31 & 71.67 & 68.27 & - & -& - &- & \textbf{67.02} & 76.45 & 71.74 \\

tqdm~\cite{pak2024textual} & EVA02-L& 68.88 & 59.18 & 70.10 & 66.05 & - &- & -&- & 64.72 & 76.15 & 70.44\\
CLOUDS~\cite{benigmim2024collaborating} & ConvNext-L &60.20 & 57.40& 67.00& 61.53 &  53.40 & 47.00 & \textbf{55.80}& 52.07 & - & - & -\\
\midrule
% \midrule
\multicolumn{13}{l}{\textit{Multi-model depth-aligned fusion}} \\

MFuser~\cite{mfuser} & DINOv2-CLIP & 71.24 & 61.08 & 71.14 & 67.82 & 53.16 & 45.59 &52.27 & 50.34 & 65.58 &  78.10& 71.84 \\
MFuser~\cite{mfuser} & DINOv2-SigLIP&  71.10 & 61.19 & 71.71 & 68.00 & 54.14* & 46.68* & 	53.57* & 	51.46* &65.44  & 77.97  &71.71 \\
% MFuser*~\cite{mfuser} & DINOv2-Depth &71.75 & 61.28& 71.73& 68.25 &54.87 & 48.37& 54.21&52.48 & 65.54&77.85 & 71.70\\
MFuser*~\cite{mfuser} & SAM-CLIP & 63.49	&58.27	& 68.10 &	63.29  &50.08	 &41.73	 &48.97	 & 46.93& 60.26&	73.16	& 66.71 \\
Vireo~\cite{chen2026leveraging} & DINOv2-Depth &70.69  & \textbf{62.91} & 69.63 &67.74 & -& -& -& -& \underline{66.73} & 75.99& 71.36\\
DepthForge~\cite{Chen_2025_ICCV} & DINOv2-Depth & 69.04 & \underline{62.82} & 69.22 &67.03 & 51.06*	 & 45.87*	& 47.89*	& 48.27* & 66.20 &76.00 & 71.10\\
DepthForge~\cite{Chen_2025_ICCV} & EVA02-Depth & 68.00 & 61.70 & 67.50 & 65.73 & - &- & -&- &62.00 & 73.00 &67.50  \\
\midrule

\multicolumn{13}{l}{\textit{Multi-model cross-depth fusion (ours)}} \\
Ours & DINOv2-CLIP & 72.37 &	62.31&	\textbf{73.52}&	\textbf{69.40}  & \textbf{55.09}	& \textbf{48.64}	& \underline{55.23}	& \textbf{52.99} & 66.03	& \textbf{78.98} &	\textbf{72.51}\\
Ours & DINOv2-SigLIP & \underline{72.79}	&62.03	& \underline{73.07}	& 69.30  	& \underline{54.51}	 &47.69	 & 54.51 & 52.24 & 65.43	& \underline{78.84}	&72.14 \\
Ours & SAM-CLIP & 64.78	& 58.29	 & 68.33	& 63.80 & 50.65		 &41.13		 &48.05		 & 46.61& 60.18& 	72.28	& 66.23\\
Ours & DINOv2-Depth & \textbf{73.20}	 & 61.98	& 73.00	& \underline{69.39} & 54.17	& \underline{48.25}	& 54.69 & 	\underline{52.37} &65.74		 &78.68		 & \underline{72.21} \\
\bottomrule
\end{tabular}
}
\end{table*}

\begin{figure}[t]
    \centering
    \includegraphics[width=0.9\textwidth]{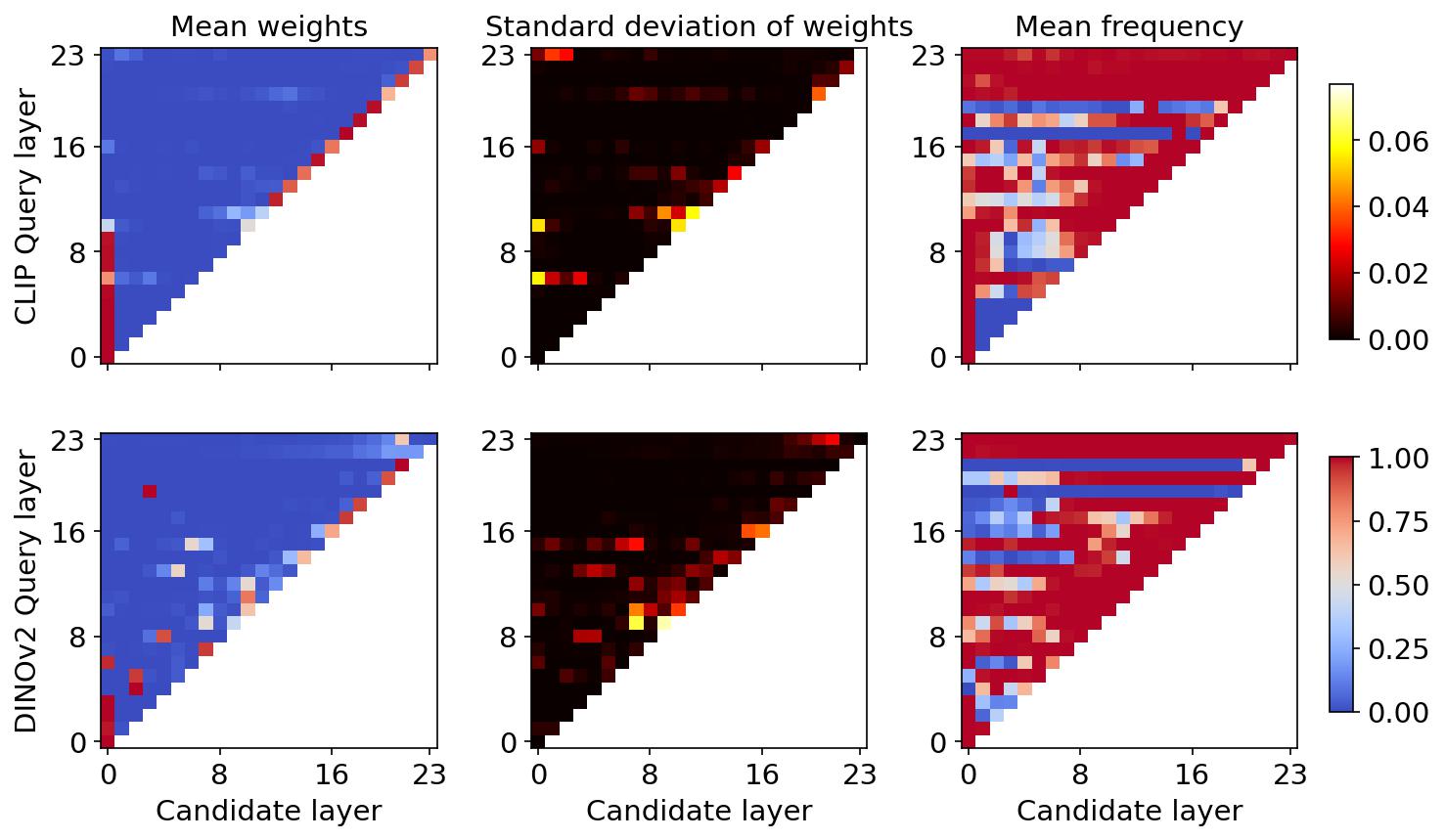}
    \caption{Layer selection statistics of SALC across domains (G, C, B, M).}
    \label{fig:layer_frequency}
\end{figure}

\subsection{In-depth layer selection analysis}
In this section, we analyse the layer selection behaviour of SALC across domains and backbones. Our analysis reveals three key findings: the learned selection (1) leverages diverse candidate layers without collapsing onto a fixed subset, (2) exhibits backbone-specific preferences that align with the complementary roles of VFM and VLM, and (3) remains largely consistent across domains. We further provide visual evidence on how candidate dropout reshapes the selection and discuss how it may benefit generalisation.  We use DINOv2-CLIP combination on the G$\rightarrow$\{C, B, M\} setting. 

\textbf{What patterns emerge in SALC's layer selection?} \quad Here, we analyse SALC's layer selection behaviour from three complementary perspectives. 

\noindent\textit{(i) Layer-wise selection patterns.} 
To gain a global perspective of selection patterns, we visualise the mean selection weights using $\mathbf{w}_t$, their standard deviations, and selection frequencies across different layers in \Cref{fig:layer_frequency}. From the mean weight,  we observe SALC develops specific patterns for different backbones. CLIP primarily leverages DINOv2's shallow layers together with the depth-aligned one, while DINOv2 prefers to draw from CLIP's middle layers. This asymmetry reflects the complementary roles of the two backbones, where DINOv2's shallow layers extract local structure that CLIP lacks, while CLIP's middle layers exhibit higher transferability~\cite{skean2025layer}. The standard deviations further show that the selection is stable across domains, with cross-domain disagreement concentrated only on a small subset of layers. Moreover, the frequency reveals that most candidates are occasionally selected, indicating that the token-wise sparsity control explores the full candidate pool rather than collapsing onto a fixed subset.

\begin{figure}[t]
    \centering
    \includegraphics[width=\linewidth]{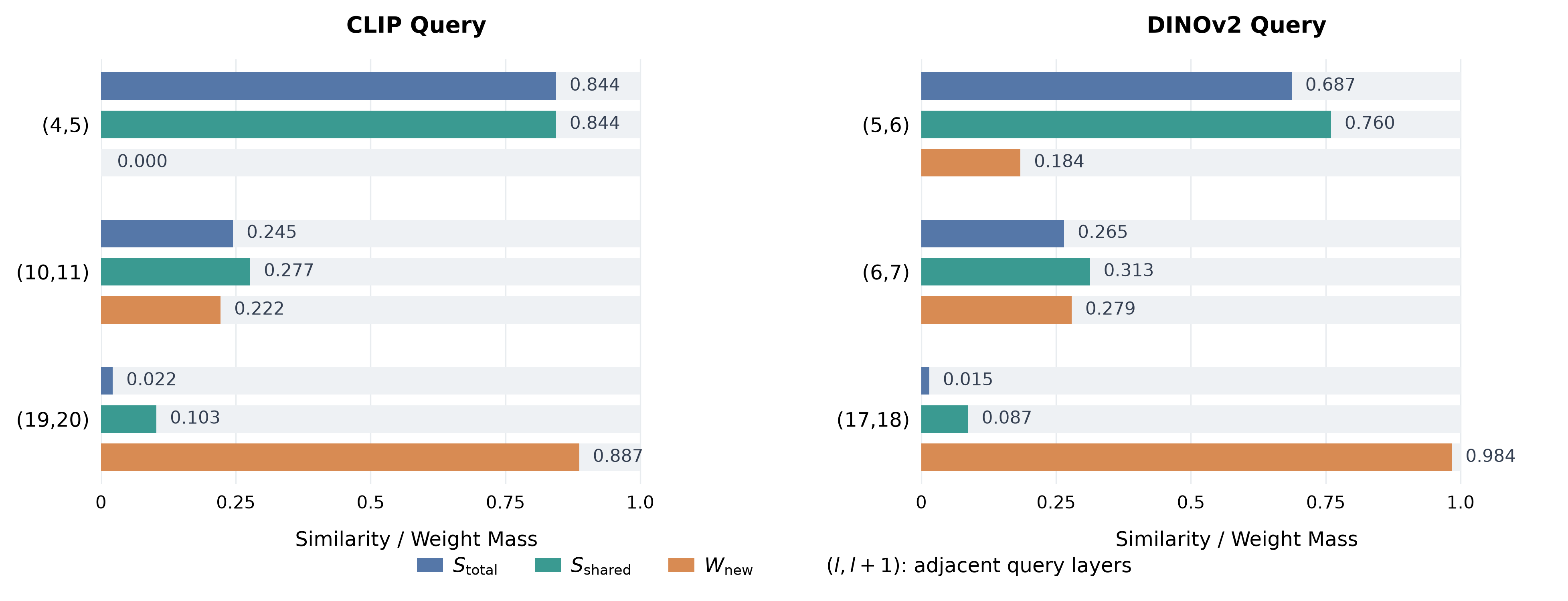}
    \caption{Representative patterns of weight redistribution across adjacent query layers.}
    \label{fig:weight_redistribution}
\end{figure}

\noindent\textit{(ii) Weight redistribution across layers.} We further investigate how selection weights evolve as new candidate layers become available. Specifically, we consider three metrics: (1) the weight distribution similarity across all candidates ($S_{\mathrm{total}}$), (2) distribution similarity across shared candidates ($S_{\mathrm{shared}}$), and (3) the weight mass assigned to the new candidate ($W_{\mathrm{new}}$). As shown in \Cref{fig:weight_redistribution}, SALC exhibits three representative patterns: (1) adjacent query layers retain similar selection preferences (first row), (2) selection weights are redistributed among existing candidates (second row), and (3) the deeper query layer prioritises the newly available candidate (third row).

\begin{figure}[t]
    \centering    
    \begin{minipage}[t]{0.48\textwidth}
        \centering
        \includegraphics[width=1\textwidth]{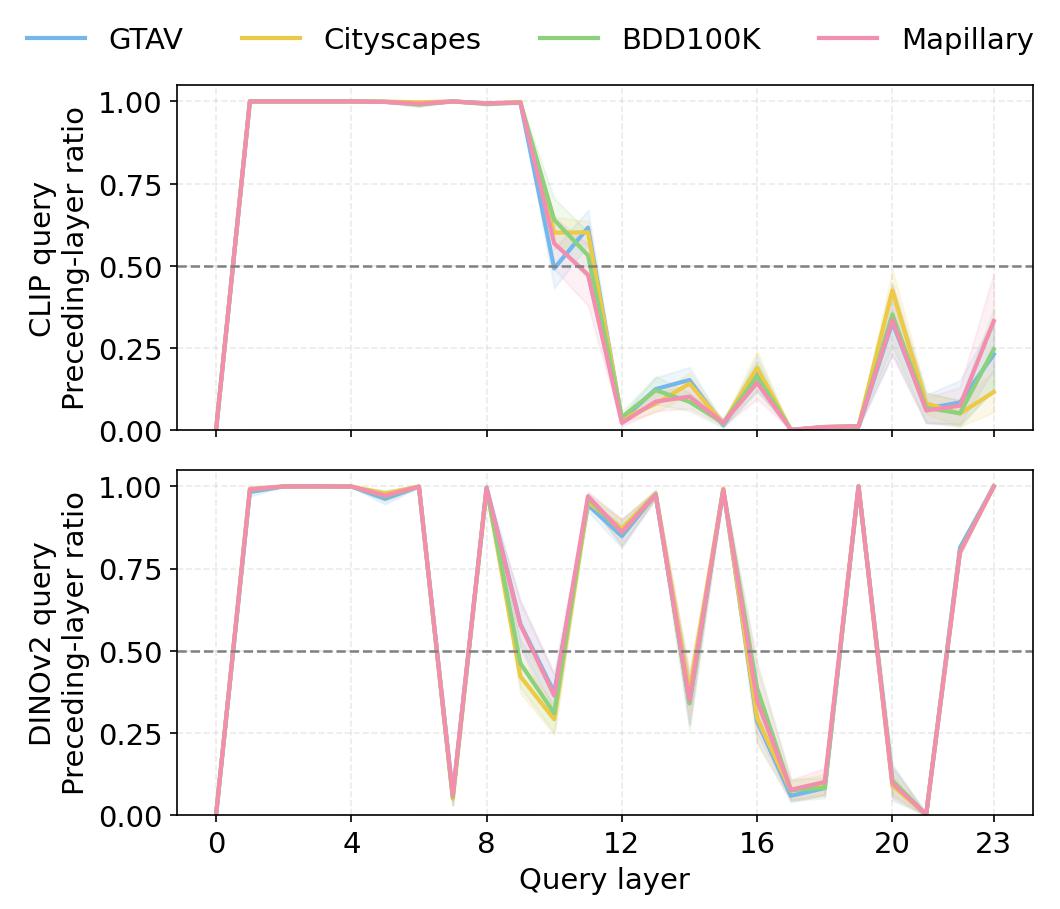}
        \caption{Preceding weight ratio across depth.}
        \label{fig:history_current}
    \end{minipage}
    \hfill
    \begin{minipage}[t]{0.48\textwidth}
        \centering
        \includegraphics[width=\textwidth]{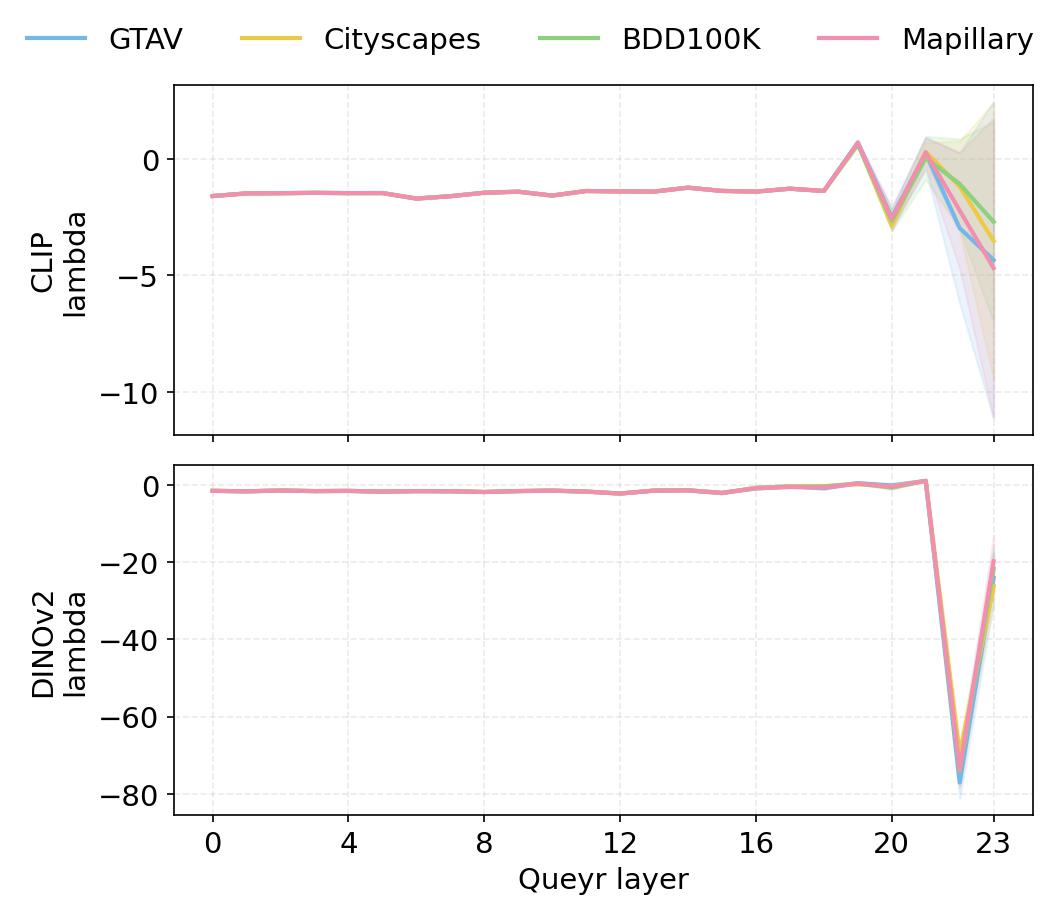}
        \caption{$\lambda$ value across layers.}
        \label{fig:lambda_dis}
    \end{minipage}
    \hfill
\end{figure}

\noindent\textit{(iii) Preference for previous vs. current layers.} We also report the ratio of preceding-layer weight to total weight in \Cref{fig:history_current}. Most query layers show a non-zero preceding ratio, confirming that SALC consistently incorporates information from preceding layers. The ratio varies substantially across depths while remaining largely consistent across domains, indicating that SALC adapts its selection at each depth rather than learning a fixed rule. Moreover, some mid and deeper query layers exhibit notable cross-domain shifts. For mid layers, the shift may stem from their inherently stronger transferability~\cite{skean2025layer, odonnat2026layer}, which allows SALC to adjust their selection more freely depending on the target domain. For deeper layers, this is likely because they have a larger candidate pool and form the final representation for decoding, which provides them more flexibility. 

\begin{figure}[t]
    \centering
        \includegraphics[width=0.9\textwidth]{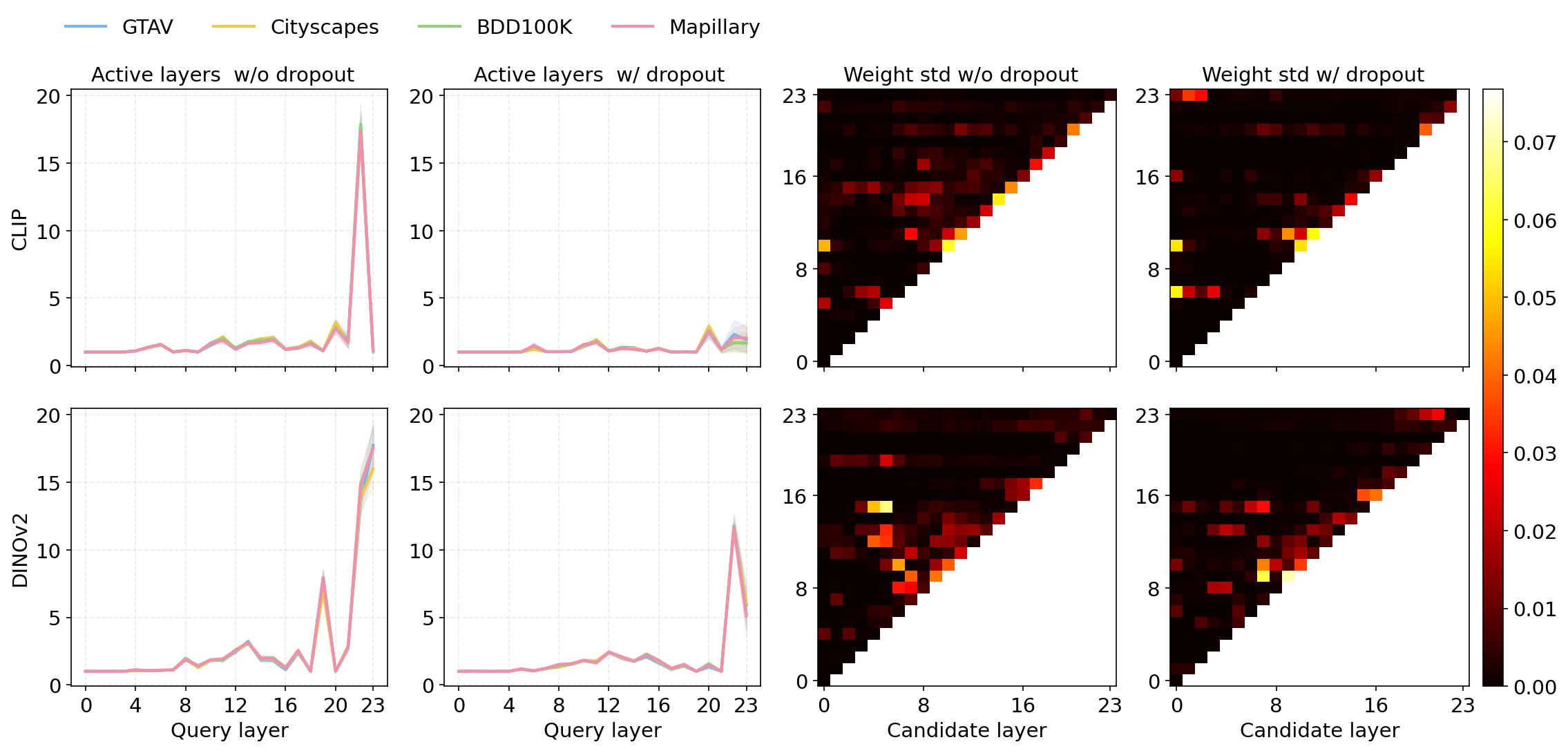}
        \caption{Active layers and cross-domain standard deviation before and after dropout.}
        \label{fig:active_layers}
\end{figure}

% frequency + pattern + ratio

\textbf{How does SALC adapt sparsity across layers and domains?} \quad
\Cref{fig:lambda_dis} shows the depth-wise distribution of the learned sparsity factor $\lambda$ across the four domains. We observe that $\lambda$ exhibits substantially higher cross-domain variability at deeper layers while remaining stable at shallow ones. We believe this is because deeper query layers tend to encode richer high-level information, which may require finer per-token sparsity adjustment to handle semantics that vary across classes and spatial locations. This pattern is also consistent with the cross-domain shift observed in the preceding ratio at deeper layers, indicating that domain variation in SALC is more visible at deeper layers through both weight assignment and sparsity control.

\textbf{How does dropout regularisation affect layer selection?} \quad
\Cref{fig:active_layers} compares the average active-layer count and cross-domain weight std before and after applying our dropout regularisation. After regularisation, the number of active candidates is clearly reduced and the selection becomes more concentrated. The cross-domain std of the selection is also reduced, as shown in the fourth column. Beyond compression, dropout exhibits distinct behaviour across query layers and models. At layers that are already moderately sparse, it does not further reduce the active-layer number but instead reorganises the dominant candidates, redirecting the selection toward different informative layers.

Based on these observations, we suggest two complementary reasons why dropout improves generalisation. First, by strengthening sparsity and lowering cross-domain variation, dropout guides  SALC to commit to the most consistently informative candidates rather than spreading weight across many candidates that may carry domain-specific noise. Second, by perturbing dominant candidates during training, dropout encourages SALC to discover diverse informative layers, leading to a more robust selection that does not depend on any specific layer.

\begin{table}[t]
\centering
\begin{minipage}[t]{0.48\linewidth}
    \centering
    \small
    \setlength{\tabcolsep}{3pt}
    \caption{Effect of masking selected layers.}
    \label{tab:layer_masking}
    \begin{tabular}{lccccc}
        \toprule
        Method & C & B & M & Avg. & $\Delta$ \\
        \midrule
        MFuser & 71.24 & 61.08 & 71.14 & 67.82 & $-1.58$ \\
        Ours & 72.37 & 62.31 & 73.52 & 69.40 & -- \\
        Mask \textit{most} & 70.79 & 59.44 & 71.11 & 67.11 & $-2.29$ \\
        Mask \textit{least} & \textbf{72.46} & \textbf{62.37} & \textbf{73.52} & \textbf{69.45} & $+0.05$ \\
        \bottomrule
    \end{tabular}%
\end{minipage}
\hfill
\begin{minipage}[t]{0.48\linewidth}
    \centering
    \small
    \setlength{\tabcolsep}{3pt}
    \caption{Extension to asymmetric architecture.}
    \label{tab:symmetric_asymmetric}
    \begin{tabular}{lccccc}
        \toprule
        Method & C & B & M & Avg. & $\Delta$ \\
        \midrule
        MFuser sym. & 71.24 & 61.08 & 71.14 & 67.82 & $-1.58$ \\
        Ours sym. & \textbf{72.37} & \textbf{62.31} & \textbf{73.52} & \textbf{69.40} & -- \\
        MFuser asym. & 67.02 & 57.67 & 69.01 & 64.56 & $-0.64$ \\
        Ours asym. & \textbf{67.04} & \textbf{58.76} & \textbf{69.81} & \textbf{65.20} & -- \\
        \bottomrule
    \end{tabular}%
\end{minipage}%
\end{table}

\textbf{Does layer selection identify critical layers?} We first use 500 images randomly sampled from the source domain to find a top-3 most-selected set of $\text{DINOv2}_\mathrm{most}=\{21, 0, 20\}$ and $\text{CLIP}_\mathrm{most}=\{21, 20, 7\}$, and a top-3 least-selected set of $\text{DINOv2}_\mathrm{least}=\{2, 1, 4\}$ and $\text{CLIP}_\mathrm{least}=\{17, 3, 16\}$. We then mask the \textit{most} and \textit{least} at inference,  respectively. As shown in \Cref{tab:layer_masking}, masking the \textit{most} set causes an obvious drop of 2.29 mIoU, falling below the depth-aligned baseline, while masking the \textit{least} leaves performance essentially unchanged. This provides evidence that the selection discriminates informative from uninformative candidates rather than distributing weight arbitrarily. 

\textbf{Can SALC be extended to models with different depths?} To improve the scalability, we can use a map function to handle two encoders $\mathcal{E}_A$ and $\mathcal{E}_B$ with different depths $L_A$ and $L_B$. Given a query layer at depth $l$ of $\mathcal{E}_A$, we map it to layer $\lfloor l \cdot \frac{L_B}{L_A}\rfloor$ of $\mathcal{E}_B$, a query layer then accesses all candidates up to its mapped depth, and vice versa. As shown in \Cref{tab:symmetric_asymmetric}, we conduct experiments using a 24-layer CLIP-Large pairing with a 12-layer DINOv2-Base (asym.) and with a 24-layer DINOv2-Large (sym.). The results show that SALC outperforms the strongest baseline by 0.64 mIoU and 1.58 mIoU.

% \textbf{Summary.}\quad 
% Together, these effects support the generalisation gains in our quantitative results.

%2x2, cross-domain, cross combo

%before after

\subsection{Ablation studies}
We conduct ablation studies to examine each component in SALC by using the DINOv2-CLIP configuration on the G$\rightarrow$\{C, B, M\} setting. More ablations are shown in Appendix~\ref{app:ablation}.

\begin{table}[t]
\begin{minipage}[t]{0.48\textwidth}
\centering
\small
\setlength{\tabcolsep}{3pt}
\caption{Ablation on sparsity control and dropout. SALC: token-wise $\lambda$ + frequency dropout.}
\label{tab:sparsity_dropout}
\begin{tabular}{lcccc}
\toprule
Variant & $\rightarrow$C & $\rightarrow$B & $\rightarrow$M & Avg. \\
\midrule
\multicolumn{5}{l}{\textit{Sparsity control}} \\
Fixed $\lambda$      & 70.62 & 60.63 & 72.27 & 67.84 \\
Layer-wise $\lambda$ & 70.67 & 60.25 & 72.61 & 67.84 \\
\midrule
\multicolumn{5}{l}{\textit{Dropout}} \\
No dropout           & 72.18	  & 61.71	   & 73.05	    & 68.98   \\
Uniform              & 71.18   & 60.13   & 72.24   & 67.85   \\
\midrule
\textbf{SALC}  & \textbf{72.37} & \textbf{62.31} & \textbf{73.52} & \textbf{69.40} \\
\bottomrule
\end{tabular}
\end{minipage}
\hfill
\begin{minipage}[t]{0.48\textwidth}
\centering
\small
\setlength{\tabcolsep}{3pt}
\caption{Ablation on candidate scope and fusion direction. SALC: all candidates + bidirectional.}
\label{tab:architecture}
\begin{tabular}{lcccc}
\toprule
Variant & $\rightarrow$C & $\rightarrow$B & $\rightarrow$M & Avg. \\
\midrule
\multicolumn{5}{l}{\textit{Candidate scope}} \\
Window size = 2            & 71.56 & 62.09 & 73.44 & 69.03 \\
Window size = 4            & 71.65 & \textbf{62.83} & 73.37 & 69.28 \\
\midrule
\multicolumn{5}{l}{\textit{Fusion direction}} \\
CLIP $\rightarrow$ DINOv2 only & 69.54   & 60.08   & 71.34   & 66.99   \\
DINOv2 $\rightarrow$ CLIP only & 71.71   & 60.68   & 73.07   & 68.49   \\
\midrule
\textbf{SALC}        & \textbf{72.37} & 62.31 & \textbf{73.52} & \textbf{69.40} \\
\bottomrule
\end{tabular}
\end{minipage}
\end{table}

\textbf{Sparsity control.} \quad
We first study the sparsity factor $\lambda$ in SALC. Specifically, we compare a fixed $\lambda$, a dynamic layer-wise $\lambda$ and our dynamic token-wise $\lambda$. The fixed $\lambda$ is set to the averaged $\lambda$ learned by SALC for a fair comparison.  As shown in \Cref{tab:sparsity_dropout}, the other two variants reduce average mIoU by 1.56. Fixed $\lambda$ removes input-dependent sparsity control, ignoring the growth of candidate layers. Layer-wise $\lambda$ loses adaptivity at the token level, which is suboptimal for dense prediction tasks. 

\textbf{Candidate dropout regularisation.} \quad We further evaluate the proposed candidate dropout regularisation by comparing uniform and frequency-based dropout. As shown in \Cref{tab:sparsity_dropout}, uniform dropout can harm generalisability because it drops candidates indiscriminately, unlike frequency-based dropout, which targets over-used candidates to strengthen sparsity and explore alternative informative layers.  We further provide multiple runs of both dropout strategies in Appendix~\ref{app:stability}, where frequency-based dropout shows smaller variance across runs.

\textbf{Dynamic selection.} \quad We also investigate whether dynamic layer selection remains necessary when the most frequently selected layers are already known. We identify the top-2 and top-4 most-selected candidates per query layer on the source domain and retrain DINOv2–CLIP with each query layer restricted to the corresponding subset.  As shown in \Cref{tab:fixed_selection}, the two variants drop by 1.81 and 2.00 mIoU, respectively. 
% This indicates that the gain of SALC does not come solely from identifying informative candidates, but also from adjusting the selection per token and per input.
While top-selected candidates carry generally robust features, restricting aggregation to them can discard necessary information retained in the remaining layers.

% We also report how dropout changes the number of active candidate layers to examine whether it encourages broader exploration of candidate layers in \Cref{fig:active_layers}.

% \begin{wraptable}{r}{0.50\textwidth}
% % \vspace{-8pt}
% \centering
% \scriptsize
% \setlength{\tabcolsep}{3pt}
% \caption{Comparison of fusion strategies.}
% \label{tab:fusion_strategies}
% \begin{tabular}{lcccc}
% \toprule
% Variant & $\rightarrow$C & $\rightarrow$B & $\rightarrow$M & Avg. \\
% \midrule
% No fusion            & 69.23 & 58.83 & 69.22 & 65.76 \\
% Depth-aligned        & 70.17	 & 60.96	& 71.38 & 67.50 \\
% Cross-depth attention& 71.85 & 60.25 & 72.37 & 68.16 \\
% Cross-depth Top2 routing & 71.66 & 59.91 & 71.73 & 67.77 \\
% \textbf{SALC}        & \textbf{72.37} & \textbf{62.31} & \textbf{73.52} & \textbf{69.40} \\
% \bottomrule
% \end{tabular}
% \end{wraptable}

\begin{table}[t]
\centering
\begin{minipage}[t]{0.48\linewidth}
    \centering
    \small
    \setlength{\tabcolsep}{3pt}
    \caption{Effect of fixed layer selection.}
    \label{tab:fixed_selection}
    % \resizebox{\linewidth}{!}{%
    \begin{tabular}{lcccc}
        \toprule
        Method & $\rightarrow$C & $\rightarrow$B & $\rightarrow$M & Avg. \\
        \midrule
        MFuser & 71.24 & 61.08 & 71.14 & 67.82\\
        Fixed Top-2 & 70.32 & 60.76 & 71.68 & 67.59 \\
        Fixed Top-4 & 70.44 & 60.67 & 71.08 & 67.40  \\
        \textbf{SALC} & \textbf{72.37} & \textbf{62.31} & \textbf{73.52} & \textbf{69.40} \\
        \bottomrule
    \end{tabular}%
    % }
\end{minipage}
\hfill
\begin{minipage}[t]{0.48\linewidth}
    \centering
    \small
    \setlength{\tabcolsep}{3pt}
    \caption{Comparison of fusion strategies.}
    \label{tab:fusion_strategies}
    % \resizebox{\linewidth}{!}{%
    \begin{tabular}{lcccc}
        \toprule
        Variant & $\rightarrow$C & $\rightarrow$B & $\rightarrow$M & Avg. \\
        \midrule
        % No fusion & 69.23 & 58.83 & 69.22 & 65.76 \\
        Depth-aligned & 70.17 & 60.96 & 71.38 & 67.50 \\
        Cross att. & 71.85 & 60.25 & 72.37 & 68.16 \\
        Cross Top2 routing & 71.66 & 59.91 & 71.73 & 67.77 \\
        \textbf{SALC} & \textbf{72.37} & \textbf{62.31} & \textbf{73.52} & \textbf{69.40} \\
        \bottomrule
    \end{tabular}%
    % }
\end{minipage}%
\end{table}

%For example, both attention and top-2 MoE struggle on BDD100K, dropping by over 2 mIoU.

% For example, two cross-depth fusion baselines struggle to handle the BDD100K domain, causing a drop of over 2 mIoU.  

\textbf{Candidate scope and fusion direction.} \quad
Beyond the core SALC mechanism, we further study the candidate scope and the cross-model fusion direction. For the candidate scope, we restrict SALC to use only the most recent 2 or 4 layers as candidate layers. For fusion direction, we disable fusion from CLIP or DINOv2 by removing SALC in their respective encoders. As reported in \Cref{tab:architecture}, reducing the candidate scope hurts performance, dropping the average mIoU by 0.37 and 0.12 for window sizes of 2 and 4, respectively. This suggests that broader access to multi-level features and bidirectional fusion benefit generalisation.  
% Moreover, removing fusion from either model degrades overall performance, indicating that bidirectional fusion is essential for SALC.

\textbf{Cross-layer fusion.} \quad
To evaluate whether our sparse all-layer fusion is more effective than existing designs, we compare four fusion strategies:
(1) depth-aligned fusion (linear connector),
(2) cross-depth fusion (attention),
(3) cross-depth fusion (Top2 routing), and
(4) cross-depth fusion (SALC). As shown in \Cref{tab:fusion_strategies}, cross-depth fusion outperforms depth-aligned fusion by leveraging multi-level features, but generalises less well than SALC on this benchmark due to the lack of dynamic sparsity.

\section{Conclusion}
In this paper, we propose a Sparse All-Layer Connector for multi-model fusion in domain generalised semantic segmentation. Unlike existing methods that fuse features only at the same depth, SALC enables each layer to leverage features from all preceding layers of the other model. To avoid redundancy that may harm generalisation, we adopt Sparsegen with a token-wise, content-dependent sparsity factor for adaptive selection across layers and tokens. In addition, we introduce a frequency-based dropout regularisation that strengthens sparsity, lowers cross-domain variation, and encourages exploration of diverse layers. We evaluate SALC on five foundation models spanning three model types, achieving competitive performance across three evaluation settings and demonstrating the effectiveness of cross-layer fusion. Beyond performance gains, our analysis further reveals that SALC learns non-trivial, backbone-specific selection patterns that adapt across depths while remaining largely consistent across domains. 

%feature image

% \section*{References}

% References follow the acknowledgments in the camera-ready paper. Use unnumbered first-level heading for
% the references. Any choice of citation style is acceptable as long as you are
% consistent. It is permissible to reduce the font size to \verb+small+ (9 point)
% when listing the references.
% Note that the Reference section does not count towards the page limit.
\clearpage
\medskip

\begingroup
  \small
  \bibliographystyle{plain}
  \bibliography{reference}
\endgroup

% {
% \small

% [1] Alexander, J.A.\ \& Mozer, M.C.\ (1995) Template-based algorithms for
% connectionist rule extraction. In G.\ Tesauro, D.S.\ Touretzky and T.K.\ Leen
% (eds.), {\it Advances in Neural Information Processing Systems 7},
% pp.\ 609--616. Cambridge, MA: MIT Press.

% [2] Bower, J.M.\ \& Beeman, D.\ (1995) {\it The Book of GENESIS: Exploring
%   Realistic Neural Models with the GEneral NEural SImulation System.}  New York:
% TELOS/Springer--Verlag.

% [3] Hasselmo, M.E., Schnell, E.\ \& Barkai, E.\ (1995) Dynamics of learning and
% recall at excitatory recurrent synapses and cholinergic modulation in rat
% hippocampal region CA3. {\it Journal of Neuroscience} {\bf 15}(7):5249-5262.
% }

%%%%%%%%%%%%%%%%%%%%%%%%%%%%%%%%%%%%%%%%%%%%%%%%%%%%%%%%%%%%
\clearpage
\appendix

\section{Active layer selection}
\label{app:sparsegen}
% \subsection{Active layer selection}
Given importance scores $\mathbf{z}_t \in \mathbb{R}^{l}$ over the $l$ 
candidate layers, Sparsegen determines the active layers by solving the 
following closed-form transformation:
\begin{equation}
\mathbf{w}_t = \underset{\mathbf{w}_t\in\mathbb{R}^l}{\arg\min}\; 
\|\mathbf{w}_t - \mathbf{z}_t\|_2^2 - \lambda_t \|\mathbf{w}_t\|_2^2,  
\quad \text{s.t. } \mathbf{w}_t \ge 0,\; \mathbf{1}^\top \mathbf{w}_t = 1,\; \lambda_t<1.
\end{equation}

\begin{proposition}[Closed-form Sparsegen, Proposition 0.1 in~\cite{laha2018controllable}]
\label{prop:sparsegen}
Let $\mathbf{z}_t \in \mathbb{R}^{l}$ be the importance scores, and let 
$z_{t,(1)} \geq z_{t,(2)} \geq \cdots \geq z_{t,(l)}$ denote the sorted 
coordinates of $\mathbf{z}_t$. Then, for any sparsity factor 
$\lambda_t \in (-\infty, 1)$, the Sparsegen solution is given by: 
\begin{equation}
w_{t,i} = \left[\frac{z_{t,i} - \tau_t}{1 - \lambda_t}\right]_+, 
\quad \forall i \in [l],
\end{equation}
where $[x]_+ = \max(x, 0)$, and the threshold $\tau_t$ is determined as
\begin{equation}
\tau_t = \frac{\sum_{i=1}^{j} z_{t,(i)} - 1 + \lambda_t}{j}, 
\quad j = \max\left\{j \in [l] \;\Big|\; 
1 - \lambda_t + j\, z_{t,(j)} > \sum_{i=1}^{j} z_{t,(i)}\right\},
\end{equation}
such that $\mathbf{w}_t$ lies on the probability simplex, 
i.e., $\sum_{i=1}^{l} w_{t,i} = 1$.
\end{proposition}

From the above closed-form solution, the active layer set $\mathcal{S}_t$ 
corresponds to the top-$j$ largest importance scores, with $|\mathcal{S}_t| = j$.

% \subsection{Computing the support set $\mathcal{S}$}

\section{Implementation details}
\label{app:details}

We provide the implementation details in \Cref{tab:app:impl}. In general, we directly use the official source code to train each framework. Moreover, we follow prior work~\cite{mfuser,chen2026leveraging,Chen_2025_ICCV} to use DINOv2-L, CLIP-L, SigLIP-L, and DepthAnythingV2-L with 24 transformer layers. We also resize their patch embeddings to a $16\times 16$ patch size to handle $512 \times 512$ inputs. We train our model on a single NVIDIA RTX 6000 Ada GPU with 48GB of memory. We also provide parameter analysis in \Cref{tab:app:params}.

\begin{table}[b]
\centering
\scriptsize
\setlength{\tabcolsep}{4pt}
\caption{Comparison of parameter counts across different methods.}
\label{tab:app:params}
\begin{tabular}{lccccccc}
\toprule
Method & Ours & Vireo~\cite{chen2026leveraging} & DepthForge~\cite{Chen_2025_ICCV} & MFuser~\cite{mfuser}  & Rein~\cite{rein} & FADA~\cite{bi2024learning} &  SoMA~\cite{yun2025soma} \\
\midrule
Trainable params (M) & 9.59  & 3.78 & 2.99 & 141.79  & 2.99 & 11.65 &  4.9 \\
\bottomrule
\end{tabular}
\end{table}

\begin{table}[t]
\centering
\caption{Implementation details of SALC integrated into the MFuser~\cite{mfuser} and DepthForge~\cite{Chen_2025_ICCV} frameworks. We follow the original training settings of each framework unless otherwise specified. SALC-specific hyperparameters are shared across both configurations.}
\label{tab:app:impl}
\begin{tabular}{lcc}
\toprule
Hyperparameter & MFuser-based & DepthForge-based \\
\midrule
\multicolumn{3}{l}{\textit{Encoder configurations}} \\
$\mathcal{E}_A$ & DINOv2/SAM & DINOv2 \\
$\mathcal{E}_B$ & CLIP / SigLIP & DepthAnythingV2 \\
% PEFT method & Rein~\cite{rein} & As in DepthForge \\
Decoder & Mask2Former & Mask2Former \\
\midrule
\multicolumn{3}{l}{\textit{Training}} \\
Optimizer & AdamW & AdamW \\
Learning rate & 1e-4 & 1e-4 \\
Weight decay & 1e-4 & 0.05 \\
LR schedule & Poly & OneCycleLR \\
Warmup iterations  & 1,500 & 4,000 \\
Total iterations & 20,000 & 40,000 \\
Batch size & 2 & 5 \\
Input resolution & 512 $\times$ 512 & 512 $\times$ 512 \\
% \midrule
% \multicolumn{3}{l}{\textit{Data augmentation}} \\
% Random crop & 1024 $\times$ 1024 & X $\times$ X \\
% Random horizontal flip & 0.5 & 0.5 \\
% Color jitter & As in MFuser & As in DepthForge \\
\midrule
\multicolumn{3}{l}{\textit{SALC-specific (shared)}} \\
% Number of SALC layers inserted & \multicolumn{2}{c}{X} \\
EMA momentum & \multicolumn{2}{c}{0.99} \\
Penalty sharpness $\gamma$ & \multicolumn{2}{c}{2} \\
Maximum drop rate $p_{\max}$ & \multicolumn{2}{c}{0.4} \\
% RMSNorm in importance scoring & \multicolumn{2}{c}{Yes} \\
\bottomrule
\end{tabular}
\end{table}

% Architecture specifications
% Training schedule
% Hyperparameter values

% \section{Additional comparisons}
% \label{app:compare}
% ACDC

\begin{table}[t]
\centering
\scriptsize
\setlength{\tabcolsep}{4pt}
\caption{Ablation study of dropout strategies under different backbone combinations on the G$\rightarrow$\{C, B, M\} setting.}
\label{tab:ablation_dropout}
\resizebox{\textwidth}{!}{
\begin{tabular}{lcccccccc}
\toprule
\multirow{2}{*}{Variant}
& \multicolumn{4}{c}{DINOv2-CLIP}
& \multicolumn{4}{c}{DINOv2-SigLIP} \\
\cmidrule(lr){2-5}
\cmidrule(lr){6-9}
& $\rightarrow$C & $\rightarrow$B & $\rightarrow$M & Avg.
& $\rightarrow$C & $\rightarrow$B & $\rightarrow$M & Avg. \\
\midrule
No dropout 
& 71.81$\pm$0.52  & 61.47$\pm$0.34 & 73.41$\pm$0.50 & 68.90$\pm$0.12
& 72.46$\pm$0.04 & 61.85$\pm$0.23 & 72.55$\pm$0.15 & 68.95$\pm$0.14 \\

Uniform dropout 
& 72.01$\pm$0.76 & 61.23$\pm$0.95 & 73.20$\pm$1.05 & 68.81$\pm$0.89
& 71.89$\pm$0.55 & 61.53$\pm$0.06 &  \textbf{73.00$\pm$0.37} & 68.81$\pm$0.30 \\

Frequency-based dropout (Ours)
& \textbf{72.03$\pm$0.36} & \textbf{62.39$\pm$0.40} & \textbf{73.44$\pm$0.08} & \textbf{69.29$\pm$0.11}
& \textbf{72.46$\pm$0.35} & \textbf{62.27$\pm$0.24} & 72.95$\pm$0.16 & \textbf{69.23$\pm$0.07} \\
\bottomrule
\end{tabular}
}
\end{table}

\section{Stability of dropout}
\label{app:stability}

Since dropout introduces stochasticity during training, we further evaluate the stability of \emph{uniform} and \emph{frequency-based} dropout by reporting the mean and standard deviation of three independent runs in \Cref{tab:ablation_dropout}. Across both DINOv2-CLIP and DINOv2-SigLIP, frequency-based dropout achieves the highest average mIoU with the smallest standard deviation ($\leq$ 0.11), while uniform dropout exhibits substantially larger variance (0.89 on DINOv2-CLIP and 0.30 on DINOv2-SigLIP). The instability of uniform dropout aligns with its design: it drops candidates indiscriminately, providing no consistent regularisation target across runs. In contrast, frequency-based dropout focuses regularisation on over-used candidates, yielding both stronger and more stable improvements.

% \section{Complexity}
% \label{app:stability}

\section{Visualisation of layer selection}
\label{app:visual}
We provide additional visualisations of the layer selection on DINOv2-SigLIP under the G$\rightarrow$
\{C, B, M\} setting. The selection patterns are consistent with those of DINOv2-CLIP shown in the main paper, confirming that SALC's behaviour generalises across backbone combinations.

\begin{figure}[t]
    \centering
    \includegraphics[width=0.9\textwidth]{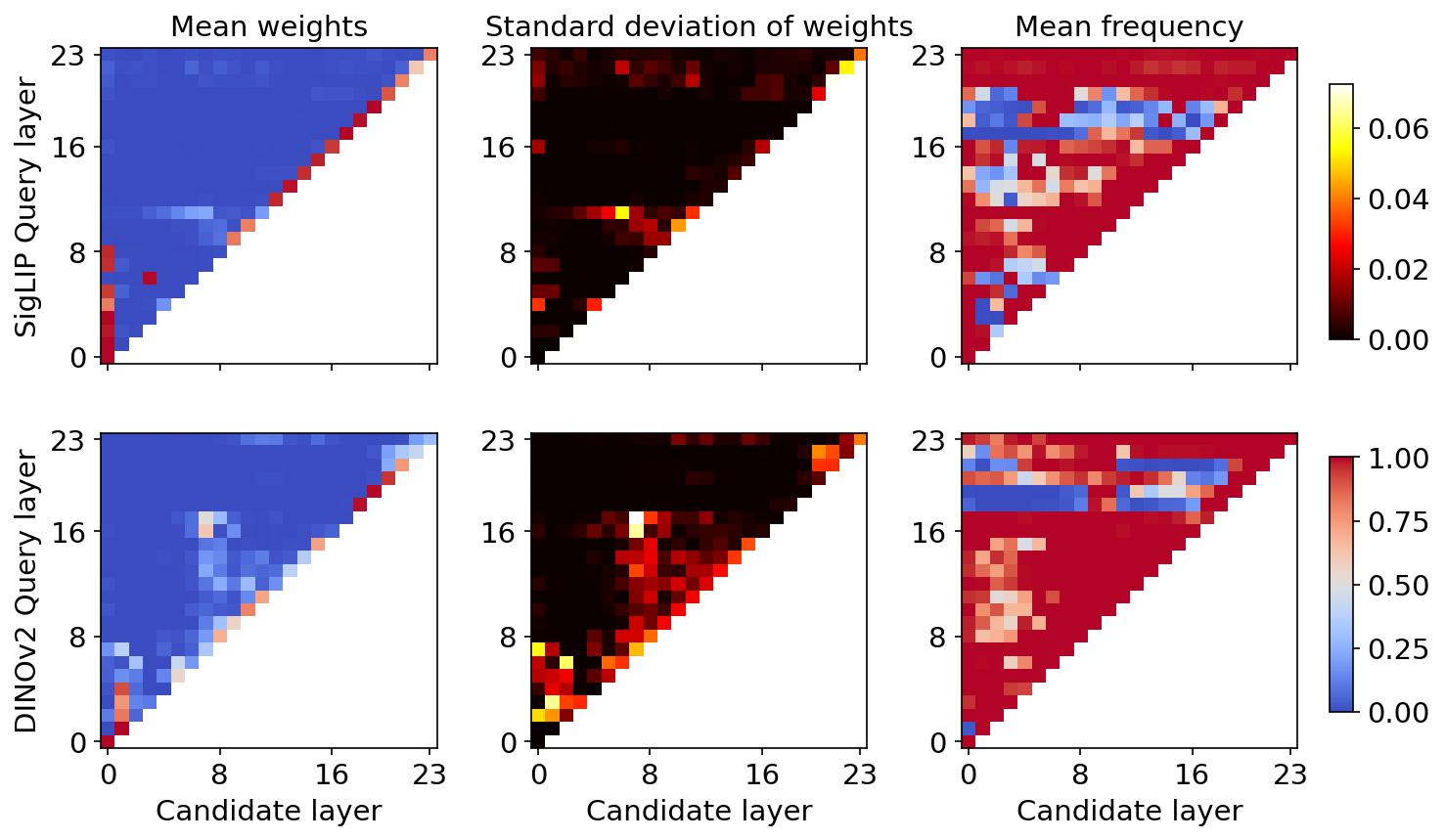}
    \caption{Layer selection statistics of SALC across domains (G, C, B, M) for DINOv2-SigLIP.}
    \label{app:fig:layer_frequency}
\end{figure}
\begin{figure}[t]
    \centering    
    \begin{minipage}[t]{0.48\textwidth}
        \centering
        \includegraphics[width=1\textwidth]{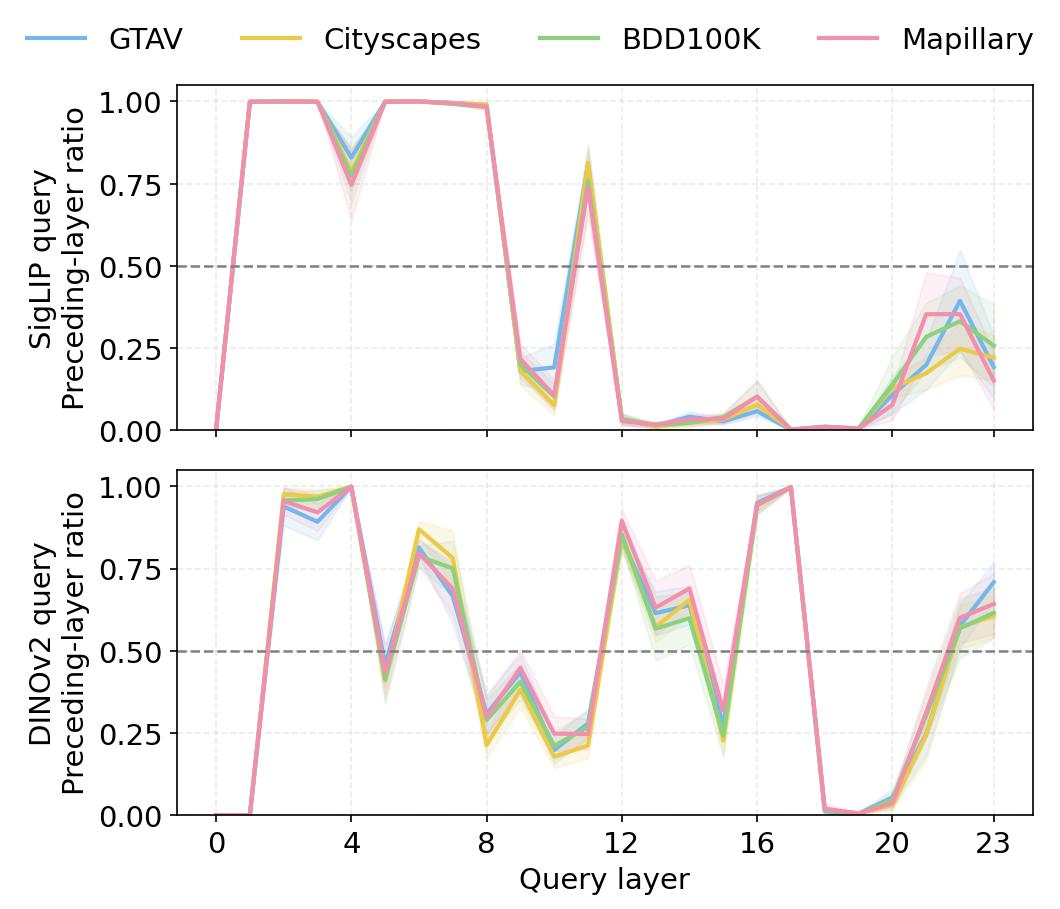}
        \caption{Preceding weight ratio across depth for DINOv2-SigLIP.}
        \label{app:fig:history_current}
    \end{minipage}
    \hfill
    \begin{minipage}[t]{0.48\textwidth}
        \centering
        \includegraphics[width=\textwidth]{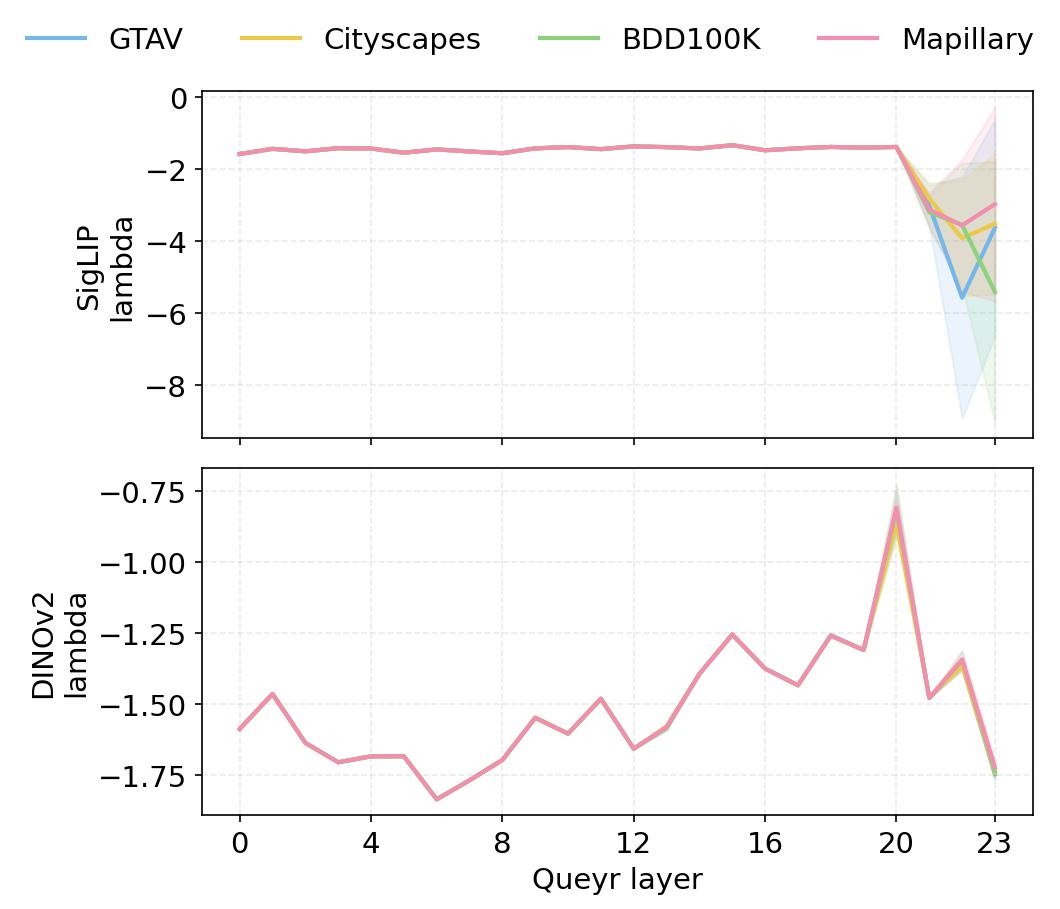}
        \caption{$\lambda$ value across layers for DINOv2-SigLIP.}
        \label{app:fig:lambda_dis}
    \end{minipage}
    \hfill
\end{figure}

\begin{figure}[t]
    \centering
        \includegraphics[width=0.9\textwidth]{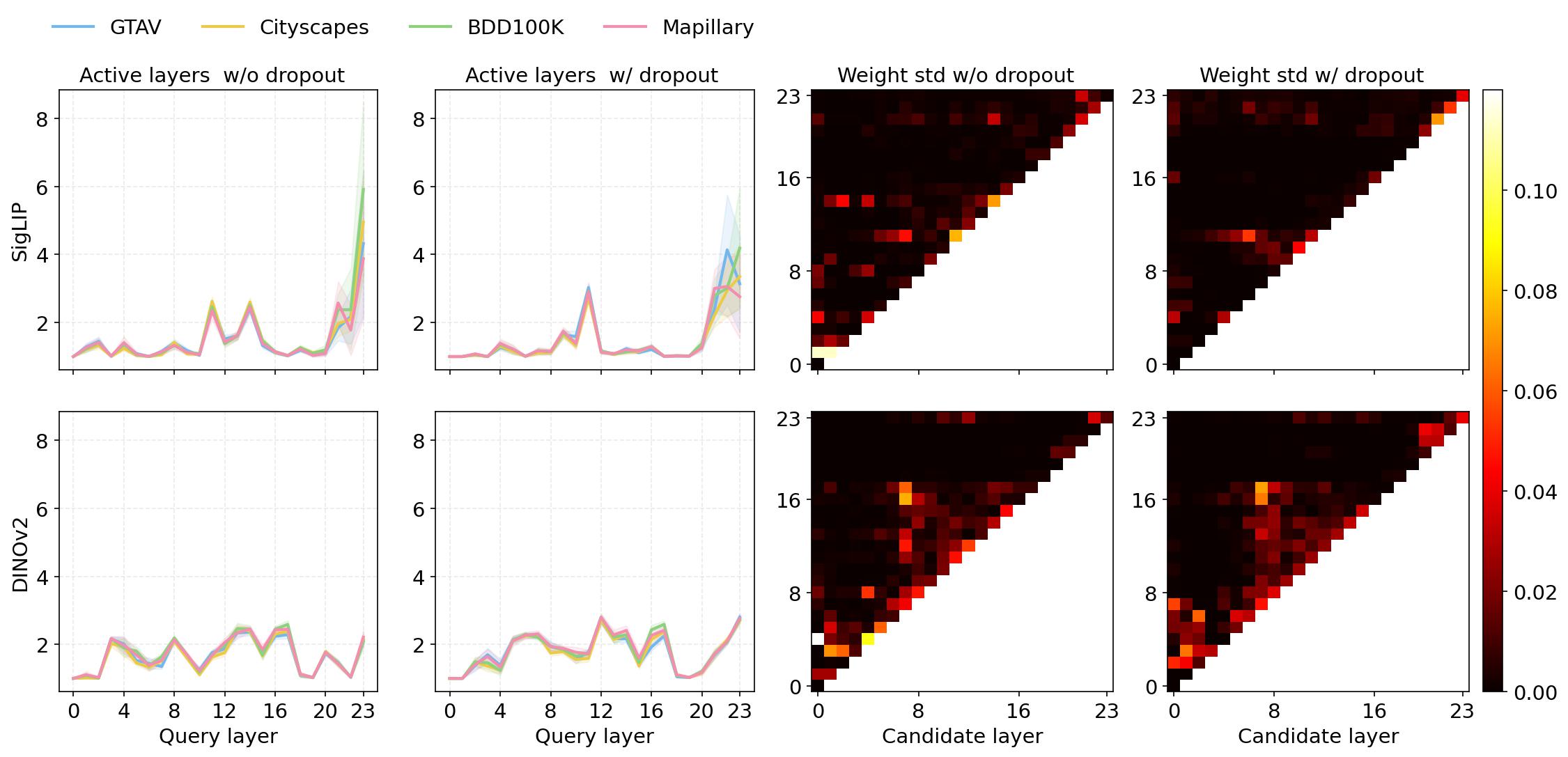}
        \caption{Active layers and cross-domain standard deviation before and after dropout for DINOv2-SigLIP.}
        \label{app:fig:active_layers}
\end{figure}

% \section{Stability of dropout}
% \label{app:stability}

\begin{figure}[t]
    \centering
    \begin{subfigure}[t]{0.48\linewidth}
        \centering
        \includegraphics[width=\linewidth]{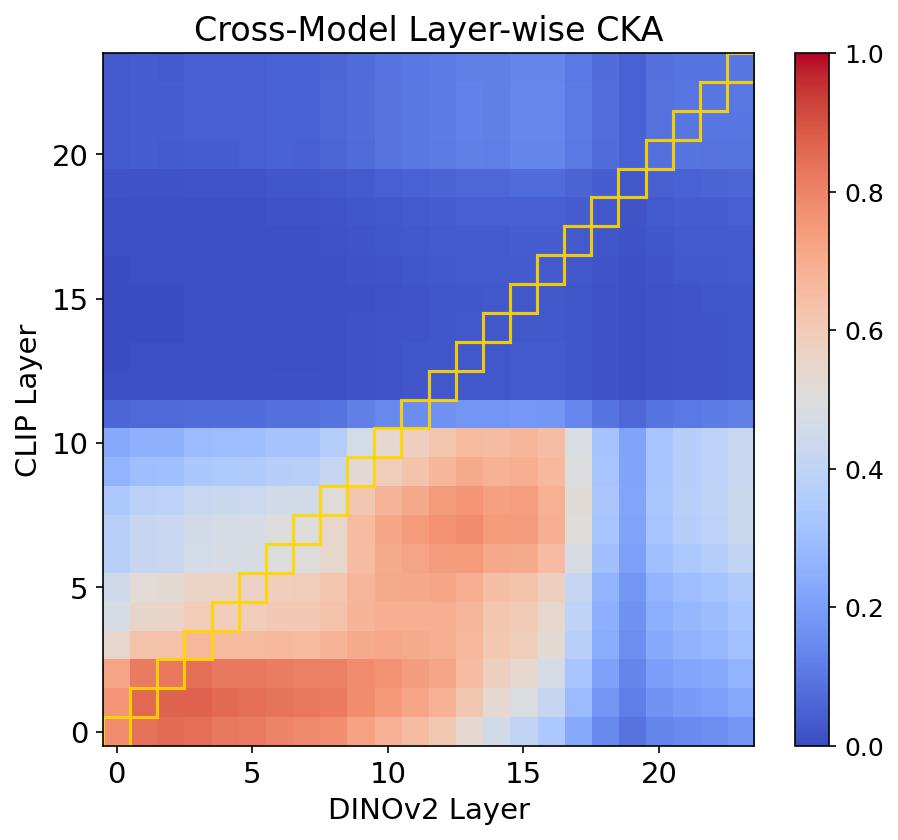}
        % \caption{CLIP vs. DINOv2}
        % \label{fig:cka_clip_dinov2}
    \end{subfigure}
    \hfill
    \begin{subfigure}[t]{0.48\linewidth}
        \centering
        \includegraphics[width=\linewidth]{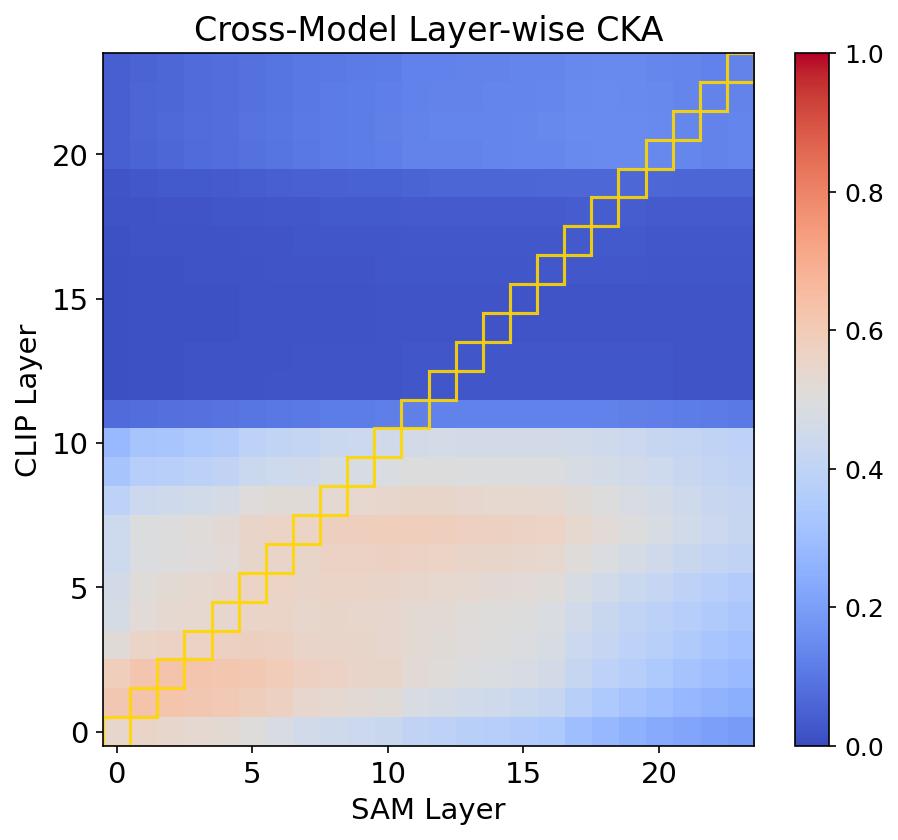}
        % \caption{CLIP vs. SAM}
        % \label{fig:cka_clip_sam}
    \end{subfigure}
    \caption{Cross-model layer-wise CKA between CLIP and different vision foundation models.}
    \label{app:cross_layer_cka}
\end{figure}

\section{Limitations}
\label{app:limit}
\subsection{Limited improvement on SAM-CLIP}
We hypothesise that the smaller gain of SALC on SAM-CLIP stems from the weak layer-wise structure of SAM features when paired with CLIP. As shown in \Cref{app:cross_layer_cka}, we compare the layer-wise Centered Kernel Alignment (CKA)~\cite{cka1} between CLIP vs DINOv2 and CLIP vs SAM. CLIP vs DINOv2 exhibits a clear layer-dependent structure: shallow CLIP layers align strongly with shallow-to-mid DINOv2 layers, while deeper CLIP layers form a distinct, less similar region. This indicates that each DINOv2 layer carries distinguishable information, providing useful signals for SALC's selection. In contrast, CLIP vs SAM shows uniformly low similarity across nearly all layer pairs, with no comparable layer-wise structure. Such weak and uniform alignment makes it difficult for SALC to identify informative SAM layers for cross-model fusion. We attribute this to SAM being trained with a different objective (segmentation supervision rather than self-distillation as in DINOv2), which leads to features less compatible with CLIP's representation space.

\subsection{Limited gain on real-to-real benchmark} 
Compared to the G$\rightarrow$\{C, B, M\} and S$\rightarrow$\{C, B, M\} settings, SALC's improvement on C$\rightarrow$
\{B, M\} is smaller, with most of the gain concentrated on Mapillary. We attribute this to the smaller domain gap between real-world source and target domains, where low-level features from the source domain transfer effectively to the targets. SALC selects layers at the semantic level rather than preserving low-level features, which limits its advantage in this setting.

\section{Additional ablations}
\label{app:ablation}
We provide additional ablations of SALC. Specifically, we study the choice 
of $\gamma$, the effect of EMA momentum, and the effect of applying RMSNorm 
before scoring. All experiments use the DINOv2--CLIP combination and are 
evaluated on the G$\rightarrow$\{C, B, M\} setting.

The EMA momentum is used to track the running statistics for our 
\emph{frequency-based dropout}, mitigating the noise introduced by the small 
batch size of 2. RMSNorm is applied to reduce the magnitude variations 
across depths in $\mathbf{F}_t$ before computing the importance scores 
$\mathbf{z}_t$. The hyperparameter $\gamma$ controls the sharpness of the 
drop-rate penalty: with $\gamma > 1$, the penalty grows more sharply for 
over-used layers.

As reported in \Cref{tab:app:ablation}, EMA improves overall generalisation 
by stabilising the dropout regularisation, while RMSNorm enhances the 
effectiveness of importance scoring across candidate layers. For $\gamma$, we observe that a mildly larger penalty on frequently selected layers 
benefits generalisation, whereas an excessively large penalty becomes 
detrimental.

\begin{table}[t]
\centering
\caption{Ablation studies on EMA, RMSNorm, and $\gamma$ using DINOv2-CLIP under the G$\rightarrow$\{C, B, M\} setting. SALC uses EMA, RMSNorm and $\gamma=2$.}
\label{tab:app:ablation}
\begin{tabular}{lcccc}
\toprule
Variant & $\rightarrow$C & $\rightarrow$B & $\rightarrow$M & Avg. \\
\midrule
\multicolumn{5}{l}{\textit{EMA}} \\
w/o EMA              &72.18  & 	61.71 &	73.05 & 68.98    \\
\midrule
\multicolumn{5}{l}{\textit{RMSNorm}} \\
w/o RMSNorm          & 72.05 & 	61.81 & 73.38 &69.08    \\
\midrule
\multicolumn{5}{l}{\textit{Penalty sharpness $\gamma$}} \\
$\gamma = 1$  & 71.69	 & 61.97	 & 73.52   & 	69.06   \\
$\gamma = 3$  & 72.16	 & 61.62	  & 73.50	 & 69.09   \\
\midrule
\textbf{SALC}  & \textbf{72.37} & \textbf{62.31} & \textbf{73.52} & \textbf{69.40} \\
\bottomrule
\end{tabular}
\end{table}

\section{Broader impacts}
\label{app:broader}
Our work studies domain generalisation in semantic segmentation, with potential applications in autonomous driving, robotics, and urban scene understanding. By improving model robustness across diverse, unseen environments, our method may help systems perform more reliably under real-world conditions such as adverse weather, nighttime, or unfamiliar scenes.
However, segmentation models may still produce unreliable predictions on rare or out-of-distribution scenes. Deploying such systems requires careful evaluation, robust failure handling, and complementary safety mechanisms. Our work does not introduce additional risks beyond those already associated with semantic segmentation in general, and all experiments use publicly available academic datasets.
% \label{app:limit}

% Sensitivity to $\rho$ (target usage)
% Sensitivity to $\gamma$ (penalty sharpness)
% Effect of EMA momentum
% RMSNorm in importance scoring
% Number of SALC layers inserted

% Technical appendices with additional results, figures, graphs, and proofs may be submitted with the paper submission before the full submission deadline (see above). You can upload a ZIP file for videos or code, but do not upload a separate PDF file for the appendix. There is no page limit for the technical appendices. 

% Note: Think of the appendix as ``optional reading'' for reviewers. The paper must be able to stand alone without the appendix; for example, adding critical experiments that support the main claims to an appendix is inappropriate. 

%%%%%%%%%%%%%%%%%%%%%%%%%%%%%%%%%%%%%%%%%%%%%%%%%%%%%%%%%%%%

\clearpage
\newpage
\section*{NeurIPS Paper Checklist}

%%% BEGIN INSTRUCTIONS %%%
The checklist is designed to encourage best practices for responsible machine learning research, addressing issues of reproducibility, transparency, research ethics, and societal impact. Do not remove the checklist: {\bf The papers not including the checklist will be desk rejected.} The checklist should follow the references and follow the (optional) supplemental material.  The checklist does NOT count towards the page
limit. 

Please read the checklist guidelines carefully for information on how to answer these questions. For each question in the checklist:
\begin{itemize}
    \item You should answer \answerYes{}, \answerNo{}, or \answerNA{}.
    \item \answerNA{} means either that the question is Not Applicable for that particular paper or the relevant information is Not Available.
    \item Please provide a short (1--2 sentence) justification right after your answer (even for \answerNA). 
   % \item {\bf The papers not including the checklist will be desk rejected.}
\end{itemize}

{\bf The checklist answers are an integral part of your paper submission.} They are visible to the reviewers, area chairs, senior area chairs, and ethics reviewers. You will also be asked to include it (after eventual revisions) with the final version of your paper, and its final version will be published with the paper.

The reviewers of your paper will be asked to use the checklist as one of the factors in their evaluation. While \answerYes{} is generally preferable to \answerNo{}, it is perfectly acceptable to answer \answerNo{} provided a proper justification is given (e.g., error bars are not reported because it would be too computationally expensive'' or ``we were unable to find the license for the dataset we used''). In general, answering \answerNo{} or \answerNA{} is not grounds for rejection. While the questions are phrased in a binary way, we acknowledge that the true answer is often more nuanced, so please just use your best judgment and write a justification to elaborate. All supporting evidence can appear either in the main paper or the supplemental material, provided in appendix. If you answer \answerYes{} to a question, in the justification please point to the section(s) where related material for the question can be found.

% IMPORTANT, please:
% \begin{itemize}
%     \item {\bf Delete this instruction block, but keep the section heading ``NeurIPS Paper Checklist"},
%     \item  {\bf Keep the checklist subsection headings, questions/answers and guidelines below.}
%     \item {\bf Do not modify the questions and only use the provided macros for your answers}.
% \end{itemize} 

%%% END INSTRUCTIONS %%%

\begin{enumerate}

\item {\bf Claims}
    \item[] Question: Do the main claims made in the abstract and introduction accurately reflect the paper's contributions and scope?
    \item[] Answer: \answerYes{} % Replace by \answerYes{}, \answerNo{}, or \answerNA{}.
    \item[] Justification: We study the multi-model fusion in domain generalised semantic segmentation.

\item {\bf Limitations}
    \item[] Question: Does the paper discuss the limitations of the work performed by the authors?
    \item[] Answer: \answerYes{} % Replace by \answerYes{}, \answerNo{}, or \answerNA{}.
    \item[] Justification: We discuss our limitations in the appendix.

\item {\bf Theory assumptions and proofs}
    \item[] Question: For each theoretical result, does the paper provide the full set of assumptions and a complete (and correct) proof?
    \item[] Answer:  \answerNA{} % Replace by \answerYes{}, \answerNo{}, or \answerNA{}.
    \item[] Justification: This paper does not include theoretical contributions.

    \item {\bf Experimental result reproducibility}
    \item[] Question: Does the paper fully disclose all the information needed to reproduce the main experimental results of the paper to the extent that it affects the main claims and/or conclusions of the paper (regardless of whether the code and data are provided or not)?
    \item[] Answer: \answerYes{} % Replace by \answerYes{}, \answerNo{}, or \answerNA{}.
    \item[] Justification: Our method is built upon the official public code of prior work. We have released our code. All training details can be found in the open public code of prior work.

\item {\bf Open access to data and code}
    \item[] Question: Does the paper provide open access to the data and code, with sufficient instructions to faithfully reproduce the main experimental results, as described in supplemental material?
    \item[] Answer: \answerYes{} % Replace by \answerYes{}, \answerNo{}, or \answerNA{}.
    \item[] Justification: We have released our code.

\item {\bf Experimental setting/details}
    \item[] Question: Does the paper specify all the training and test details (e.g., data splits, hyperparameters, how they were chosen, type of optimizer) necessary to understand the results?
    \item[] Answer: \answerYes{} % Replace by \answerYes{}, \answerNo{}, or \answerNA{}.
    \item[] Justification: We provide training details in the main paper and appendix. We use the default setting of prior work.

\item {\bf Experiment statistical significance}
    \item[] Question: Does the paper report error bars suitably and correctly defined or other appropriate information about the statistical significance of the experiments?
    \item[] Answer: \answerNo{} % Replace by \answerYes{}, \answerNo{}, or \answerNA{}.
    \item[] Justification: For dropout regularisation, we report results over multiple runs. In addition, our motivation experiments are conducted with 40 random seeds.

\item {\bf Experiments compute resources}
    \item[] Question: For each experiment, does the paper provide sufficient information on the computer resources (type of compute workers, memory, time of execution) needed to reproduce the experiments?
    \item[] Answer: \answerYes{} % Replace by \answerYes{}, \answerNo{}, or \answerNA{}.
    \item[] Justification: We provide training details in the appendix.
    
\item {\bf Code of ethics}
    \item[] Question: Does the research conducted in the paper conform, in every respect, with the NeurIPS Code of Ethics \url{https://neurips.cc/public/EthicsGuidelines}?
    \item[] Answer: \answerYes{} % Replace by \answerYes{}, \answerNo{}, or \answerNA{}.
    \item[] Justification: The research conducted in this paper conforms to the NeurIPS Code of Ethics.

\item {\bf Broader impacts}
    \item[] Question: Does the paper discuss both potential positive societal impacts and negative societal impacts of the work performed?
    \item[] Answer: \answerYes{} % Replace by \answerYes{}, \answerNo{}, or \answerNA{}.
    \item[] Justification: We discuss our broader impacts in the appendix.
    
\item {\bf Safeguards}
    \item[] Question: Does the paper describe safeguards that have been put in place for responsible release of data or models that have a high risk for misuse (e.g., pre-trained language models, image generators, or scraped datasets)?
    \item[] Answer: \answerNA{} % Replace by \answerYes{}, \answerNo{}, or \answerNA{}.
    \item[] Justification: The paper does not involve the release of data or models with a high risk of misuse.

\item {\bf Licenses for existing assets}
    \item[] Question: Are the creators or original owners of assets (e.g., code, data, models), used in the paper, properly credited and are the license and terms of use explicitly mentioned and properly respected?
    \item[] Answer: \answerYes{} % Replace by \answerYes{}, \answerNo{}, or \answerNA{}.
    \item[] Justification: The creators of the code, datasets, and pretrained models used in this work are properly credited through citations. We use publicly available assets in accordance with their respective licenses and terms of use.

\item {\bf New assets}
    \item[] Question: Are new assets introduced in the paper well documented and is the documentation provided alongside the assets?
    \item[] Answer: \answerYes{} % Replace by \answerYes{}, \answerNo{}, or \answerNA{}.
    \item[] Justification: We have released the code upon acceptance and provide appropriate documentation alongside it.

\item {\bf Crowdsourcing and research with human subjects}
    \item[] Question: For crowdsourcing experiments and research with human subjects, does the paper include the full text of instructions given to participants and screenshots, if applicable, as well as details about compensation (if any)? 
    \item[] Answer: \answerNA{} % Replace by \answerYes{}, \answerNo{}, or \answerNA{}.
    \item[] Justification: This paper does not involve crowdsourcing experiments or research with human subjects.

\item {\bf Institutional review board (IRB) approvals or equivalent for research with human subjects}
    \item[] Question: Does the paper describe potential risks incurred by study participants, whether such risks were disclosed to the subjects, and whether Institutional Review Board (IRB) approvals (or an equivalent approval/review based on the requirements of your country or institution) were obtained?
    \item[] Answer: \answerNA{} % Replace by \answerYes{}, \answerNo{}, or \answerNA{}.
    \item[] Justification: This paper does not involve research with human subjects, so IRB approval and participant risk disclosure are not applicable.

\item {\bf Declaration of LLM usage}
    \item[] Question: Does the paper describe the usage of LLMs if it is an important, original, or non-standard component of the core methods in this research? Note that if the LLM is used only for writing, editing, or formatting purposes and does \emph{not} impact the core methodology, scientific rigor, or originality of the research, declaration is not required.
    %this research? 
    \item[] Answer: \answerNA{} % Replace by \answerYes{}, \answerNo{}, or \answerNA{}.
    \item[] Justification: LLM is only used for editing and formatting.

\end{enumerate}

\end{document}